\documentclass[12pt]{article}

\usepackage[left=1in, right=1in, top=1in, bottom=1in]{geometry}

\usepackage[bitstream-charter]{mathdesign}
\usepackage{amsmath}
\usepackage[scaled=0.92]{PTSans}
\usepackage{xfrac}

\usepackage{titling}

\usepackage[table,usenames,dvipsnames]{xcolor}
\definecolor{shadecolor}{gray}{0.9}

\usepackage[english]{babel}
\usepackage[parfill]{parskip}
\usepackage{afterpage}
\usepackage{framed}
\usepackage{verbatim}
\usepackage{setspace}

{\endMakeFramed}

\usepackage{lineno}

\usepackage{ragged2e}

\newcounter{parcount}

\usepackage[symbol]{footmisc}

\usepackage[inline]{enumitem}

\usepackage{graphicx}
\usepackage[labelfont={it, small}, font=small]{caption}
\usepackage[format=hang]{subcaption}
\usepackage{wrapfig}

\usepackage{chngcntr}

\usepackage{booktabs}
\usepackage{longtable}
\usepackage{hhline}

\usepackage[algoruled]{algorithm2e}
\SetKwInput{KwSample}{sample}
\SetKwInput{KwDefine}{define}
\SetKwInput{KwSet}{set}
\usepackage{listings}
\usepackage{fancyvrb}
\fvset{fontsize=\normalsize}

\usepackage{amsthm}
\newtheorem{proposition}{Proposition}
\newtheorem{lemma}{Lemma}

\usepackage[numbers,compress]{natbib}

\usepackage[colorlinks,linktoc=all]{hyperref}
\hypersetup{citecolor=MidnightBlue}
\hypersetup{linkcolor=black}
\hypersetup{urlcolor=MidnightBlue}

\usepackage[nameinlink]{cleveref}

\usepackage[acronym,nowarn]{glossaries}

\lstdefinestyle{mystyle}{
    commentstyle=\color{OliveGreen},
    keywordstyle=\color{BurntOrange},
    numberstyle=\tiny\color{black!60},
    stringstyle=\color{MidnightBlue},
    basicstyle=\ttfamily,
    breakatwhitespace=false,
    breaklines=true,
    captionpos=b,
    keepspaces=true,
    numbers=left,
    numbersep=5pt,
    showspaces=false,
    showstringspaces=false,
    showtabs=false,
    tabsize=2
}
\usepackage[colorinlistoftodos,
            prependcaption,
            textsize=small,
            backgroundcolor=yellow,
            linecolor=lightgray,
            bordercolor=lightgray]{todonotes}

\usepackage{mdframed}
\newmdenv[
  topline=false,
  bottomline=false,
  rightline=false,
  leftline=false,
  skipabove=\topsep,
  backgroundcolor=black!10,
  ]{siderules}
  
\newcounter{example}[section]

\crefname{figure}{Figure}{Figures}
\Crefname{figure}{Figure}{Figures}

\newtheorem{theorem}{Theorem}
\newtheorem{assumption}{Assumption}

\newtheorem{corollary}[theorem]{Corollary}

\usepackage{soul}

\renewcommand{\mid}{~\vert~}

\newcommand{\loc}{m}
\newcommand{\weight}{w}
\newcommand{\params}{\theta}
\newcommand{\latentrate}{\nu}

\newcommand{\dif}{\mathop{}\!\mathrm{d}}

\newcommand{\E}{\mathbb{E}}

\newcommand{\bbI}{\mathbb{I}}

\newcommand{\cA}{\mathcal{A}}

\newcommand{\cC}{\mathcal{C}}

\newcommand{\cN}{\mathcal{N}}

\newcommand{\cX}{\mathcal{X}}
\newcommand{\cY}{\mathcal{Y}}

\newcommand{\naturals}{\mathbb{N}}
\newcommand{\reals}{\mathbb{R}}

\newcommand{\distCategorical}{\mathrm{Cat}}
\newcommand{\distDirichlet}{\mathrm{Dir}}
\newcommand{\distGamma}{\mathrm{Ga}}
\newcommand{\distInvWishart}{\mathrm{IW}}

\newcommand{\distNegBinomial}{\mathrm{NB}}
\newcommand{\distNormal}{\mathcal{N}}
\newcommand{\distPoisson}{\mathrm{Po}}

\newcommand{\distPoissonProcess}{\mathrm{PP}}
\newcommand{\distUniform}{\mathrm{Unif}}

\RequirePackage{collcell}
\RequirePackage{eqparbox}

\newacronym[longplural={Neyman-Scott processes}]{NSP}{NSP}{Neyman-Scott process}
\newacronym{DPMM}{DPMM}{Dirichlet process mixture model}
\newacronym{MFMM}{MFMM}{mixture of finite mixture model}
\newacronym{RJMCMC}{RJMCMC}{reversible jump Markov chain Monte Carlo}
\newacronym{INLA}{INLA}{integrated nested Laplace approximation}
\newacronym{GMM}{GMM}{Gaussian mixture model}
\newacronym{EPPF}{EPPF}{exchangeable partition probability function}
\newacronym{KL}{KL}{Kullback-Leibler}
\newacronym{ELBO}{ELBO}{\emph{evidence lower bound}}
\newacronym{EM}{EM}{\emph{expectation-maximization}}
\newacronym{SVI}{SVI}{stochastic variational inference}

\usepackage{titlesec}

\titleformat{\section}
  {\normalfont\fontsize{13}{13}\bfseries}{\thesection}{1em}{}
\titleformat{\subsection}
  {\normalfont\fontsize{12}{13}\bfseries}{\thesubsection}{1em}{}

\allowdisplaybreaks

\begin{document}

\pretitle{\begin{center}\Large}
\posttitle{\end{center}}
\title{Spatiotemporal Clustering with Neyman-Scott Processes via Connections to Bayesian Nonparametric Mixture Models}
\preauthor{\begin{center}\large \hspace{-0.25em}}
\postauthor{\end{center}}
\author{Yixin Wang$^{1,*}$, Anthony Degleris$^{2,*}$, Alex Williams$^{3,4}$, and Scott W. Linderman$^{5,\dagger}$}
\predate{}
\postdate{\vspace{-1em}}
\date{}
\maketitle

\doublespacing
\allowdisplaybreaks

\thispagestyle{empty}

\begin{figure}[!b]
\begin{minipage}[l]{\textwidth}
  \small
  \singlespacing
  \rule{.3\linewidth}{.5pt} \\
  $^1$ Department of Statistics, University of Michigan, Ann Arbor, MI, USA \\
  $^2$ Department of Electrical Engineering, Stanford University, Stanford, CA, USA \\
  $^3$ Center for Neural Science, New York University, New York, NY, USA \\
  $^4$ Center for Computational Neuroscience, Flatiron Institute, New York, NY, USA \\
  $^5$ Department of Statistics and Wu Tsai Neurosciences Institute, Stanford University, Stanford, CA, USA \\
  $^*$ Equal Contribution \\
  $^\dagger$ Corresponding Author: \texttt{scott.linderman@stanford.edu}
\end{minipage}
\end{figure}

\begin{abstract} \noindent
\glspl{NSP} are point process models that generate clusters of points in time or space.
They are natural models for a wide range of phenomena, ranging from neural spike trains to document streams.
The clustering property is achieved via a doubly stochastic formulation: first, a set of latent events is drawn from a Poisson process; then, each latent event generates a set of observed data points according to another Poisson process. 
This construction is similar to Bayesian nonparametric mixture models like the \gls{DPMM} in that the number of latent events (i.e.\ clusters) is a random variable, but the point process formulation makes the \gls{NSP} especially well suited to modeling spatiotemporal data.
While many specialized algorithms have been developed for \glspl{DPMM}, comparatively fewer works have focused on inference in \glspl{NSP}.
Here, we present novel connections between \glspl{NSP} and \glspl{DPMM}, with the key link being a third class of Bayesian mixture models called \glspl{MFMM}.
Leveraging this connection, we adapt the standard collapsed Gibbs sampling algorithm for \glspl{DPMM} to enable scalable Bayesian inference on \gls{NSP} models. 
We demonstrate the potential of Neyman-Scott processes on a variety of applications including sequence detection in neural spike trains and event detection in document streams.
\end{abstract}

\begin{small}
\end{small}

\section{Introduction}
\label{sec:intro}

Many natural systems give rise to discrete sets of events in time and space.
Biological neurons communicate with one another via sequences of spikes.
Social network users share time-stamped messages with text, images, and other metadata.
A fundamental challenge is to develop statistical models and inference algorithms that can provide insight into such systems and enable accurate predictions of future events.

Poisson processes are clearly suited to this task~\citep{kingman1992poisson}.
However, real systems may not obey the strict independence assumptions these models impose.
In a Poisson process, the number of events in one interval of time is independent of the number of events in nearby but non-overlapping intervals. 
This assumption is often violated in practice.

Consider a simple model of a bursty neuron~\citep{krahe2004burst}.
Suppose the neuron typically fires action potentials (spikes) at a low rate of~$\lambda_0=2$ spikes/second, and in this regime it may be well modeled as a homogenous Poisson process.
However, suppose that every so often the neuron emits bursts of around $w=5$ action potentials in the span of a few hundred milliseconds.
If 3 spikes are measured in a 100ms window of time, it is likely that the neuron is in the middle of a burst so we would expect to see more spikes in the following 100ms as well.

While these dependencies across disjoint intervals defy the basic assumptions of a Poisson process, the neural spike times may still be well modeled as \textit{conditionally} Poisson.
For example, suppose we know there are~$L$ bursts that occur at times~$\{m_l\}_{l=1}^L \subset [0, T]$.
Then the spike times might be modeled as an inhomogeneous Poisson process with conditional intensity~\citep{Daley2003},
\begin{align*}
  \lambda(t \mid \{m_l\}_{l=1}^L) &= \lambda_0 + w \sum_{l=1}^L \cN(t; m_l, 0.1^2)
\end{align*}
where $\cN(t; \mu, \sigma^2)$ denotes the Gaussian density function with mean $\mu$ and variance $\sigma^2$.
According to this model, weighted Gaussian \textit{impulse responses} are superimposed on top of the constant background rate to produce bursts of spikes, clustered together in time.

\glsresetall \Glspl{NSP}~\citep{Neyman1958} are point processes that capture this clustering property.
In an \gls{NSP}, the cluster centers (the burst times in our example) are treated as latent variables.
Conditioned on these latent variables, the observed events follow an inhomogeneous Poisson process.
Marginalizing over the latent variables induces dependencies across intervals that could not be captured with a simple Poisson process.

The Neyman-Scott process follows a simple generative model.
First, a set of \textit{latent events} is drawn from a Poisson process.
Then, each latent event adds an impulse response to the intensity function of another Poisson process, which produces the set of \textit{observed events}.
If the impulse responses are localized in time and space, then each latent event produces a cluster of observed events.
The Neyman-Scott process is a type of \textit{doubly stochastic} point process (a.k.a. Cox process) since the intensity itself is a stochastic process. 

The set of underlying latent events is often of scientific interest.
For example, we may want to relate a neuron's burst times to a sensory stimulus or behavioral covariate. 
However, this can be challenging from a statistical perspective, since neither the number of latent events nor their locations are known.
Traditionally, inference and estimation in \glspl{NSP} has been approached with \gls{RJMCMC} algorithms~\citep{Green2003}, which use birth and death moves to address this trans-dimensional inference problem~\citep{Moller2003}.
However, in practice, \gls{RJMCMC} can suffer from high rejection rates, sacrificing performance unless proposals are carefully crafted.
Alternatively, there are a number of specialized algorithms based on minimum contrast estimation, which optimize the parameters of the \gls{NSP} to match statistics of the data, such as Ripley's K-function~\citep{Ripley1977-yo} or the pair correlation function~\citep{Bartlett1963-gh,Cressie1993-tq,Stoyan1996-tf,Waagepetersen2007-ar,Diggle2013-sg,Baddeley2015-mq}.
However, obtaining closed-form expressions for these second-order statistics often requires strong assumptions about the parametric form of the \gls{NSP}~\citep{Tanaka2014, Moller2014}.

Here, we establish a novel connection between \glspl{NSP} and Bayesian nonparametric mixture models, and in doing so we inherit the host of Bayesian inference algorithms that have been developed for the latter~\citep{Neal2000,Jain2004,Jain2007,Blei2006}. 
Intuitively, latent events are akin to clusters in a mixture model.
The key observation is that in an \gls{NSP}, the number of latent events is a Poisson random variable.
In this regard, the \gls{NSP} differs from the \gls{DPMM}~\citep{antoniak1974mixtures,Lo1984-ls,neal1992bayesian,Escobar1995} --- the canonical Bayesian nonparametric mixture model --- since a \gls{DPMM} produces a countably infinite number of clusters.
Instead, the \gls{NSP} more closely resembles the~\gls{MFMM}, which assumes an explicit prior distribution on the number of clusters~\citep{Miller2018,Nobile1994-am, phillips1996bayesian,Richardson1997-hy}.
In an~\gls{NSP}, that prior is a Poisson distribution and, in contrast to other \glspl{MFMM}, its mean is tuned to the volume of space or time under consideration, making the \gls{NSP} especially well-suited to spatiotemporal modeling.

In this paper, we will formalize the connection between Neyman-Scott processes and mixture of finite mixture models. 
Furthermore, we show how Dirichlet process mixture models arise as a limiting form of an \gls{NSP}, and in doing so we bridge~\glspl{MFMM} and~\glspl{DPMM}.
We exploit the novel connection to derive a collapsed Gibbs sampling algorithm like that of~\citet{Neal2000} and~\citet{Miller2018}, which operates directly on partitions of data points.
Finally, we demonstrate the efficacy of this algorithm on simulated data and in applications to sequence detection in multineuronal spike train recordings and event detection in document streams.

\section{Neyman-Scott Processes}
\label{sec:nsp}

\begin{figure}[t]
\centering
\includegraphics[width=\linewidth]{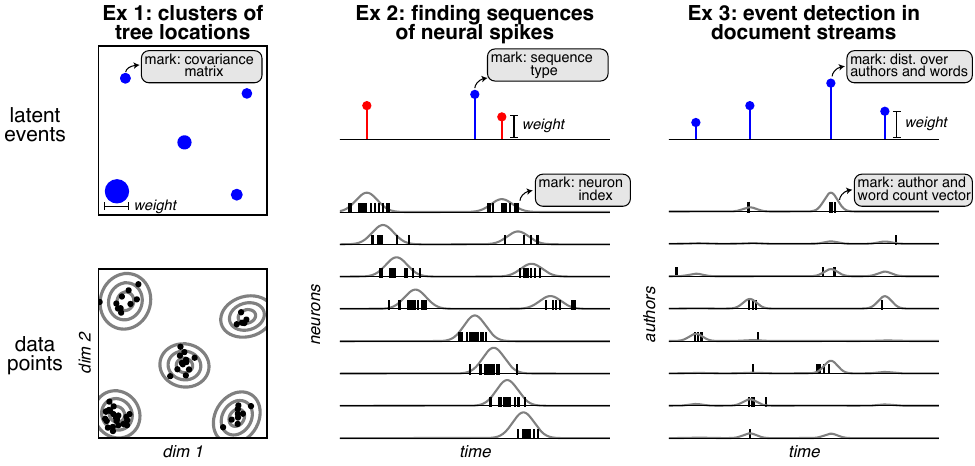}
\caption[Three examples of Neyman-Scott processes]{
\textit{Three example applications of Neyman-Scott processes.}
Example 1 (left) is a spatial clustering model with Gaussian impulse responses. Each latent event has a location, weight, and a covariance matrix as its parameter (as shown in the gray box, e.g.). The weight determines the expected number of induced data points.
Example 2 (middle) is an \gls{NSP} for finding recurring sequences of spikes in neural recordings. Each latent event has a time, weight, and type (red or blue) as its parameter. The induced data points are spikes with a time and a neuron assignment (i.e. mark). The latent event type determines which neurons participate in the sequence and at what delay.
Example 3 (right) is an \gls{NSP} for detecting events in document streams. Each latent event has a time, weight, and a parameter that specifies the distributions over authors and words. The induced data points are documents with a time, author, and a vector of word counts.
In all three examples, goals include inferring latent events, filling in missing data, and predicting data points in unobserved regions of time or space.
}
\label{fig:examples}
\end{figure}

Point processes are probabilistic models that yield random finite sets of points~$\{x_n\}_{n=1}^N \subset \cX$, where $\cX$ is a compact subset of $\reals^D$, e.g. of space or time~\citep{Daley2003}.
The simplest point process is the Poisson process~\citep{kingman1992poisson}, which is governed by a nonnegative intensity function~${\lambda(x): \cX \mapsto \reals_+}$. 
The number of events that fall in a region $\cX' \subseteq \cX$ follows a Poisson distribution with mean~${\int_{\cX'} \lambda(x) \dif x}$. 
Moreover, the number of events in $\cX'$ is independent of the number in $\cX''$ if $\cX'$ and $\cX''$ are disjoint. 
We use the notation~$\{x_n\}_{n=1}^N \sim \distPoissonProcess(\lambda(x))$ to indicate that a set~$\{x_n\}_{n=1}^N$ is drawn from a Poisson process with intensity function $\lambda(x)$. 
%
A \emph{marked} Poisson process is a Poisson process that generates sets of tuples~$\{(x_n, y_n)\}_{n=1}^N$ with locations $x_n \in \cX$ and marks $y_n \in \cY$ according to an intensity function~$\lambda(x, y)$. 



Neyman-Scott processes~\citep{Neyman1958} are \emph{doubly stochastic processes} (also known as \emph{Cox processes}~\citep{Cox1955-dh}) in which the intensity is derived from a Poisson process.
First, a set of \textit{latent events} is drawn from a marked Poisson process,
\begin{align}
    \{(\loc_l, \weight_l, \params_l)\}_{l=1}^L
    &\sim \distPoissonProcess \big(\latentrate(\loc, \weight, \params) \big).
\end{align}
Each latent event includes a location~$\loc_l \in \cX$, a non-negative weight~$\weight_l \in \reals_+$, and, optionally, extra parameters,~$\params_l \in \Theta$.
Without loss of generality, assume~$\latentrate(\loc, \weight, \params) = \overline{L}(\cX) \, p(\loc, \weight, \params)$, where~$p(\loc, \weight, \params)$ is a normalized probability density and~$\overline{L}(\cX)$ is the total measure.
Thus, the number of latent events is a Poisson random variable, $L \sim \distPoisson(\overline{L}(\cX))$.

Then, each latent event adds a non-negative \textit{impulse response} to the intensity of another marked Poisson process, which produces the observed data points,
\begin{align}
    \label{eq:nsp_intensity_1}
    \lambda(x, y \mid \{(\loc_l, \weight_l, \params_l)\}_{l=1}^L) &=
    \sum_{l=1}^L \weight_l \, p(x, y \mid \loc_l, \params_l), \\
    \label{eq:nsp_sampling}
    \{(x_n, y_n)\}_{n=1}^N &\sim \distPoissonProcess \left( \lambda(x, y \mid \{(\loc_l, \weight_l, \params_l)\}_{l=1}^L) \right).
\end{align}
As above, we assume $p(x, y \mid \loc_l, \params_l)$ is a probability density which is scaled up or down by the weight,~$w_l$.
Dependencies between the weights, locations, and parameters can be incorporated into the latent event intensity function~$\latentrate(\loc, \weight, \params)$.

\subsection{Example applications of Neyman-Scott processes}
\label{sec:examples}

Neyman-Scott processes are applicable to a wide variety of spatiotemporal clustering problems. 
\Cref{fig:examples} highlights three examples, which we will return to in our experiments. 

\paragraph{Modeling the locations of trees in a forest} Consider modeling the locations of trees in a forest.
Trees of the same species may cluster together if they grow from seeds dropped by a shared ancestor.
Neyman-Scott processes have been used to model this type of spatial point process data~\citep{stoyan2000recent,Wiegand2013-de}.
Let~$\cX \subset \reals^2$ denote a compact subset of the 2-dimensional plane, like the square shown in \Cref{fig:examples} (left).
Likewise, let $m_l$, $\theta_l$, and $w_l$  denote the location, covariance, and weight of a cluster of trees, respectively, and assume the impulse responses are scaled Gaussian densities, $w_l \cN(x_l; m_l, \theta_l)$. 
In~\Cref{fig:examples} (left), we set the latent event intensity to be homogenous in space and proportional to a gamma density and inverse Wishart density in the weight and covariance, respectively.

\paragraph{Finding sequences of spikes in neural recordings}
With multielectrode arrays, neuroscientists can record the times at which individual neurons fire action potentials, or \textit{spikes}. 
A sequence of spike times from one or more neurons is called a \textit{spike train}.
We can model a spike train as a collection of tuples~$\{(x_n, y_n)\}_{n=1}^N$ where~$x_n \in \cX = [0,T]$ denotes the time of the~$n$-th spike and~${y_n \in \cY = \{1,\ldots,Y\}}$ denotes the index of the neuron on which it occurred. 
An important problem in neuroscience is detecting sequences of spikes that unfold across multiple neurons and recur many times over the course of a recording~\citep{Abeles2001,Russo2017}.
This is an extension of the burst-detection problem in Section~\ref{sec:intro}.

Consider the following Neyman-Scott process model for sequence detection, as described by~\citet{Williams2020_ppseq}. 
Each sequence corresponds to a latent event, which consists of a time (i.e. location)~$\loc_l \in [0,T]$, weight~$\weight_l \in \reals_+$, and a discrete type (i.e. parameter) $\params_l \in\{1, \ldots, S\}$. 
The hyperparameter $S$ specifies the number of sequence types, and each neuron may spike with different probability and delay under different sequence types.
By varying the offsets, a single latent event can induce a sequence of spikes across a set of neurons, and different sequence types can engage different subsets of neurons, as shown in Figure~\ref{fig:examples} (middle).


\paragraph{Detecting world events from streams of news documents}
Finally, suppose we have a collection of documents, each with a timestamp, author, and a string of words.
For example, they may be diplomatic cables exchanged between multiple countries. 
We represent the dataset as~$\{(x_n, y_n)\}_{n=1}^N$ where~$x_n \in \cX = [0,T]$ denotes the time stamp.
The mark is another tuple~$y_n = (y_n^{(a)}, y_n^{(c)})$, where~$y_n^{(a)} \in \{1,\ldots,A\}$ denotes the index of the document's author and~$y_n^{(c)} \in \naturals_0^V$ denotes a vector of word counts for each term in a vocabulary of size~$V$.
Compared to previous examples, these are very high dimensional marks.

The idea is that documents that discuss the same event were likely written around the same time and use similar terms. 
For example, we might see a flurry of diplomatic cables following an election.
In many cases, those events are latent and we aim to infer them from the observed documents.
To perform event detection with a Neyman-Scott process, let the latent events of the~\gls{NSP} consist of a time~$\loc_l \in [0,T]$, weight~$\weight_l \in \reals_+$, and parameters~$\params_l = (\params_l^{(a)}, \params_l^{(c)})$.  
The first parameter is a distribution over authors; the second is a vector of intensities for each word in the vocabulary, as illustrated in~\Cref{fig:examples} (right).

\subsection{Simulating a Neyman-Scott process}
\label{sec:simulating}
These examples bear strong resemblance to mixture models, with each latent event corresponding to a cluster.
The \textit{Poisson superposition principle}~\citep{kingman1992poisson} makes this connection explicit.
Given the set of latent events, the intensity function in \cref{eq:nsp_intensity_1} is a sum of non-negative impulse responses. 
To simulate such a Poisson process, we can independently sample data points from Poisson processes for each impulse response and take their union.

To sample data points for each impulse response, first sample the number of induced data points, then sample their locations and marks independently according to the normalized impulse response.  
Since the impulse responses are weighted probability density functions, the expected number of induced data points is simply the weight, $w_l$,
and the normalized impulse responses are simply the conditional densities,~$p(x, y \mid \loc_l, \params_l)$.
This procedure for sampling an \gls{NSP} is shown in~\Cref{alg:sample_nsp1}.

This sampling procedure lets us interpret the Neyman-Scott process as a clustering model in which each data-point is attributed to of one of the underlying latent events.
Moreover, under the~\gls{NSP}, the number of clusters~$L$ is a random variable, suggesting a connection between Neyman-Scott processes and Bayesian nonparametric mixture models. 
We formalize this relationship in the next section.

\begin{figure}[t]
  \centering
    \begin{minipage}[t]{0.49\textwidth}
    \begin{algorithm}[H]
    \KwSample{$L \sim \distPoisson(\overline{L}(\cX))$}
    \For{$l=1,\ldots,L$}{
    \KwSample{$\loc_l, \weight_l, \params_l \sim p(\loc, \weight, \params)$} 
    \KwSample{$N_l \sim \distPoisson(\weight_l)$} 
    \For{$n=1,\ldots,N_l$}{
        \KwSample{\\\hspace{1em}$x_{l, n}, y_{l, n} \sim p(x, y \mid \loc_l, \params_l)$}
        }
    }
    \Return $\cup_{l=1}^L \{(x_{l, n}, y_{l, n})\}_{n=1}^{N_l}$
    \caption{Sample an \gls{NSP}}
    \label{alg:sample_nsp1}
    \end{algorithm}
    \end{minipage}
    \hfill
    \begin{minipage}[t]{0.49\textwidth}
    \begin{algorithm}[H]
        \KwSample{$L \sim \distPoisson(\overline{L}(\cX))$}
        \For{$l=1,\ldots,L$}{
        \KwSample{$\loc_l, \weight_l, \params_l \sim p(\loc, \weight, \params)$}
        }    
        \KwSet{$W = \sum_{l=1}^L \weight_l$}
        \KwSet{$\pi = \left(\sfrac{\weight_1}{W}, \ldots, \sfrac{\weight_L}{W} \right)$}
        \KwSample{$N \sim \distPoisson(W)$}
        \For{$n=1,\ldots,N$} {
        \KwSample{$z_n \sim \distCategorical(\pi)$}
        \KwSample{$x_n, y_n \sim p(x, y \mid \loc_{z_n}, \params_{z_n})$}
        }
        \Return $\{(x_{n}, y_n)\}_{n=1}^{N}$
        \caption{Sample an \gls{NSP} (v2)}
        \label{alg:sample_nsp2}
    \end{algorithm}
    \end{minipage}
\label{fig:sample_nsp}
\end{figure}

\section{Neyman-Scott Processes and Bayesian Nonparametric Mixture Models}
\glsreset{DPMM}
\glsreset{MFMM}

Our first contribution is to establish a connection between Neyman-Scott processes, \glspl{DPMM} \citep{antoniak1974mixtures,Lo1984-ls,neal1992bayesian,Escobar1995}, and the \gls{MFMM} recently studied by \citet{Miller2018}. 
\Glspl{MFMM} are Bayesian mixture models in which the number of clusters is a finite random variable.
We will show how the NSP can be seen as an \gls{MFMM} in which the distribution of the number of components depends on how many data points are observed. 
This relationship allows us to extend many of the results of \citet{Miller2018} and characterize the \gls{NSP} in terms of its random partition distribution, urn scheme, and random measure formulation. 
In doing so, we will see that the Neyman-Scott Process occupies a privileged place among \glspl{MFMM}: in a certain limit, the \gls{NSP} reduces to the classic \gls{DPMM}. 
This relationship allows the Neyman-Scott process to interpolate between these two important model classes and inherit the efficient inference algorithms available for \glspl{DPMM} and \glspl{MFMM}.

To establish the connection to \glspl{MFMM} and \glspl{DPMM}, we first present an equivalent sampling procedure for Neyman-Scott processes.
Rather than sampling the numbers of induced data points, $N_l$, for each of the latent events, we first sample the total number of induced data points, $N$, and then assign each data point a \textit{parent} latent event.
Since each $N_l$ is an independent Poisson random variable, the total number of induced data points is also Poisson distributed and the parents are categorical random variables~\citep{kingman1992poisson}.
The alternative algorithm is shown in~\Cref{alg:sample_nsp2}.

\begin{figure}[t]
  \centering
    \begin{minipage}[t]{0.49\textwidth}
    \begin{algorithm}[H]
        \KwSample{$L \sim \distPoisson(\overline{L}(\cX))$}
        \For{$l=1,\ldots,L$}{
        \KwSample{$\loc_l, \params_l \sim p(\loc, \params)$}
        }
        \KwSample{$W \sim \distGamma(L\alpha, \beta)$}
        \KwSample{$\pi \sim \distDirichlet(\alpha 1_l)$}
        \KwSample{$N \sim \distPoisson(W)$}
        \For{$n=1,\ldots,N$} {
        \KwSample{$z_n \sim \distCategorical(\pi)$}
        \KwSample{$x_n, y_n \sim p(x, y \mid \loc_{z_n}, \params_{z_n})$}
        }
        \Return $\{(x_{n}, y_n)\}_{n=1}^{N}$
        \caption{Sample an \gls{NSP} with gamma weights}
        \label{alg:sample_nsp3}
    \end{algorithm}
    \end{minipage}
    \hfill
    \begin{minipage}[t]{0.49\textwidth}
    \begin{algorithm}[H]
        \KwSample{$L \sim \distPoisson(\overline{L}(\cX))$}
        \For{$l=1,\ldots,L$}{
        \KwSample{$\loc_l, \params_l \sim p(\loc, \params)$}
        }
        \KwSample{$\pi \sim \distDirichlet(\alpha 1_L)$}
        \KwSample{$N \sim \distNegBinomial(L\alpha, (1+\beta)^{-1})$}
        \For{$n=1,\ldots,N$} {
        \KwSample{$z_n \sim \distCategorical(\pi)$}
        \KwSample{$x_n, y_n \sim p(x, y \mid \loc_{z_n}, \params_{z_n})$}
        }
        \Return $\{(x_{n}, y_n)\}_{n=1}^{N}$
        \caption{Sample an \gls{NSP} with gamma weights (v2)}
        \label{alg:sample_nsp4}
    \end{algorithm}
    \end{minipage}
\label{fig:sample_nsp2}
\end{figure}

The sampling procedure simplifies even further under the following assumption.
\begin{assumption}
\label{asmp:gamma_weights}
The latent event intensity factors as,
\begin{align}
    \latentrate(\loc, \weight, \params) &= \overline{L}(\cX) \, \distGamma(\weight \mid \alpha, \beta) \, p(\loc, \params). 
\end{align}
That is, the latent event weights are gamma random variables with shape~$\alpha$ and inverse scale~$\beta$, and they are independent of the latent event locations and parameters. 
\end{assumption}
Under this assumption, the Neyman-Scott process becomes a \textit{shot noise G Cox process}~\citep{Brix1999-if,Moller2003}, which is closely related to the Poisson-gamma process~\citep{Wolpert1998-tg}, multiplicative intensity models~\citep{ishwaran2004computational}, and L{\'e}vy-moving averages~\citep{james2005bayesian}. 
We will simply call it an~\gls{NSP} with gamma weights.

The total weight in~\Cref{alg:sample_nsp2} is the sum of independent and identically distributed gamma random variables, which is also a gamma distributed,~$W \sim \distGamma(L\alpha, \beta)$.
Moreover, the normalized weights in \Cref{alg:sample_nsp2} follow a symmetric Dirichlet distribution,~$\pi \sim \distDirichlet(\alpha 1_L)$, where~$1_L$ is a length-$L$ vector of ones. 
Under the gamma distribution, the total weight is independent of the normalized weights~\citep{lukacs1955characterization}.
\Cref{alg:sample_nsp3} uses these three properties to sample an~\gls{NSP} with gamma weights, per Assumption~\ref{asmp:gamma_weights} above.
Finally, since the gamma and Poisson are conjugate, we can marginalize over the total weight to obtain a negative binomial distribution on the number of data points, as in collapsed sampling procedure in \Cref{alg:sample_nsp4}.

Compare~\Cref{alg:sample_nsp4} to \Cref{alg:sample_mfmm} for sampling an \gls{MFMM}, adapted from~\citet{Miller2018}.
There are three main differences.  
First, the \gls{MFMM} allows for arbitrary distributions on the number of components,~$p_L$, a p.m.f.~on the positive integers, whereas the \gls{NSP} has a Poisson distribution.
Second, the number of latent events in an \gls{NSP} depends on the total measure over the domain~$\cX$; that is, an \gls{NSP} can generalize from one domain to another and scale the expected number of latent events accordingly.
Third, the \gls{MFMM} treats the number of observed events $N$ as fixed, whereas the \gls{NSP} model treats $N$ as a random variable that depends on the number of latent events,~$L$.

Conditioned on the number of data points, however, the Neyman-Scott process with gamma weights is a special case of the mixture of finite mixtures model. 
In an~\gls{NSP}, the number of data points carries information about the number of latent events---intuitively, more data points suggests more clusters.
In particular, given that we have observed $N$ data points, the \gls{NSP} with gamma weights is a special case of the \gls{MFMM} in which~$p(L \mid N) \propto \distPoisson(L \mid \overline{L}(\cX)) \cdot \distNegBinomial(N \mid L \alpha, (1 + \beta)^{-1})$, where $\distNegBinomial$ is the negative binomial probability mass function (pmf).
This distribution is closely related to a shifted confluent hypergeometric distribution, which we call the \textit{Schein distribution} after~\citet{schein2019poisson}.
The Schein distribution is in turn closely related to the P\'{o}lya-Aeppli distribution~\citep{johnson2005univariate}. 
Its pmf has a closed-form expression in terms of hypergeometric functions.

\begin{wrapfigure}{r}{0.45\textwidth}
    \singlespacing
    \vspace{-2.5em}
    \begin{algorithm}[H]
        \KwSample{$L \sim p_L$ (a p.m.f.~on $\{1,2,\ldots\}$)}
        \For{$l=1,\ldots,L$}{
        \KwSample{$\loc_l, \params_l \sim p(\loc, \params)$}
        }
        \KwSample{$\pi \sim \distDirichlet(\alpha 1_L)$}
        \For{$n=1,\ldots,N$} {
            \KwSample{$z_n \sim \distCategorical(\pi)$}
            \KwSample{$x_n, y_n \sim p(x, y \mid \loc_{z_n}, \params_{z_n})$}
        }
        \Return $\{(x_{n}, y_n)\}_{n=1}^{N}$
        \caption{Sample an \gls{MFMM} with $N$ data points}
        \label{alg:sample_mfmm}
    \end{algorithm}
    \vspace{-.5em}
\end{wrapfigure}
The correspondence between \glspl{NSP} and \glspl{MFMM} enables many of the results of \citet{Miller2018} to be adapted to the Neyman-Scott process.
In particular, we will derive the marginal distribution over partitions of $N$ data points in an \gls{NSP} and a corresponding P\'{o}lya urn scheme, which leads to yet another algorithm for sampling Neyman-Scott processes.
This urn scheme will suggest an efficient collapsed Gibbs sampling algorithm for posterior inference in \glspl{NSP} with gamma weights, which we develop in \Cref{sec:inference}.
Moreover, the urn scheme reveal a novel connection between Neyman-Scott processes and Dirichlet process mixture models.

\subsection{The exchangeable partition distribution of a Neyman-Scott process}
The Neyman-Scott process induces a random partition of data points into clusters associated with different latent events. 
Let $N$ denote the number of observed data points and $\cC$ denote a partition of the indices~$\{1,\ldots,N\}$ induced by the \textit{parent assignments}~$z_n$ in Algorithm~\ref{alg:sample_nsp4}. 
In other words,~$\cC$ is a set of disjoint, non-empty sets whose union is~$[N] = \{1, \ldots, N\}$.  
We represent the partition as~$\cC = \{\cC_k: |\cC_k| > 0\}$, where~$\cC_k = \{n: z_n = k\}$ is the set of data indices assigned to latent event~$k$ and $|\cC_k|$ is the size of that cluster.
The size of the partition,~$|\cC|$, is the number of latent events that induced at least one observed data point, so we must have~$L \geq |\cC|$.


The parent assignments induce a partition of the observed events, but they are tied to a particular labeling of the latent events.  
The partition, by contrast, is invariant to permutations of the latent event indices. 
Working with partitions frees us from keeping track of latent event indices or empty components in mixture models---we only need to infer how the finite number of data points are partitioned into different components. 

\begin{theorem}
\label{thm:partition-prior}
Under Assumption~\ref{asmp:gamma_weights}, the prior probability of the partition induced by an \gls{NSP}, integrating over the latent event locations, weights, and parameters, is,
\begin{align}
\label{eq:epdist}
p(N,\cC) = V_{N,|\cC|} \prod_{\cC_k \in \cC} \frac{\Gamma(|\cC_k| + \alpha)}{\Gamma(\alpha)},
\end{align}
where $\alpha$ and $\beta$ are the shape and rate, respectively, of the gamma prior on latent event weights, $|\cC|$ is the number of clusters in the partition, $|\cC_k|$ is the size of the $k$-th cluster in the partition, $N = \sum_{\cC_k \in \cC} |\cC_k|$ is the total number of data points (a random variable), and
\begin{align}
\label{eq:Vncoef}
V_{N,|\cC|} = \frac{1}{N!} \left(\frac{1}{1+\beta} \right)^{N} \sum_{L=|\cC|}^\infty \distPoisson(L \mid \overline{L}(\cX)) \, \frac{L!}{(L-|\cC|)!} \, \left(\frac{\beta}{1+\beta}\right)^{L \alpha}.
\end{align}
\end{theorem}
Theorem~\ref{thm:partition-prior} and its proof (given in Appendix~\ref{app:proofs}) closely parallel the main theorem of~\citet{Miller2018}, which gives the partition distribution for mixture of finite mixture models. 
The main difference is that eq.~\ref{eq:epdist} is a distribution on partitions of random size~$N \in \naturals_0$.
To obtain a conditional distribution~$p(\cC \mid N)$ on partitions of given size~$N$---the partition distribution more commonly considered~\citep{pitman2006,Miller2018}---we need to divide by the probability of obtaining~$N$ data points,
\begin{align}
    p(N) &= \sum_{L=0}^\infty \distPoisson(L \mid \overline{L}(\cX)) \, \distNegBinomial(N \mid L\alpha, (1+\beta)^{-1}).
\end{align}
The distribution above is a Poisson-mixed negative binomial. 
It does not, to our knowledge, have an analytical form, but we will not need one for our purposes.

Note that~$p(\cC \mid N)$ is a symmetric function of the cluster sizes,~$|\cC_k|$, and therefore it is invariant to permutations of the integers~$[N]$.
Thus,~$\cC$ is an \emph{exchangeable} random partition and~$p(\cC \mid N)$ is an \gls{EPPF}~\citep{pitman2006}.
In particular, it takes the form of a Gibbs partition~\citep{Gnedin2006} since it factors into a term that depends on the number of data points and clusters ($\sfrac{V_{N,|C|}}{p(N)}$) and a product of rising factorials ($\sfrac{\Gamma(|\cC_k| + \alpha)}{\Gamma(\alpha)}$) depending on the size of each cluster.



The \gls{EPPF} offers yet another way of sampling a Neyman-Scott process as shown in Algorithm~\ref{alg:sample_nsp5}. 
First sample a random partition~$\cC$, which specifies the number of data points; then sample parameters for each cluster in the partition; and finally, independently sample data points for each cluster. 
The key step is sampling a random partition, and as we will see in the next section, this can be done with a simple P\'{o}lya-urn process.
The main advantage of this formulation is that it only involves the partition, cluster parameters, and data points, and that enables the simple collapsed Gibbs sampling algorithm for \glspl{NSP} presented in \Cref{sec:inference}. 

\begin{figure}[t]
  \centering
    \begin{minipage}[t]{0.46\textwidth}
    \begin{algorithm}[H]
        \KwSample{$N, \cC \sim p(N, \cC)$}
        \For{$\cC_k \in \cC$}{
            \KwSample{$\loc_k, \params_k \sim p(\loc, \params)$}
            \For{$n \in \cC_k$}{
                \KwSample{\\ \hspace{1em} $x_n, y_n \sim p(x, y \mid \loc_k, \params_k)$}
            }
        }
        
        \Return{$\{(x_{n}, y_n)\}_{n=1}^{N}$}
        \caption{Sample an \gls{NSP} with gamma weights (v3)}
        \label{alg:sample_nsp5}
    \end{algorithm}
    \end{minipage}
    \hfill
    \begin{minipage}[t]{0.53\textwidth}
    \begin{algorithm}[H]
        \KwSample{$L \sim \distPoisson(\overline{L}(\cX))$}
        \KwSample{$N \sim \distNegBinomial(L\alpha, (1+\beta)^{-1})$}
        \KwSet{$\cC_1 = \{1\}$ and $\cC = \{\cC_1\}$}
        \For{$n=2,\ldots,N$}{
        \KwSet{$Z = n-1 + |\cC|\alpha + \alpha \overline{L}(\cX) \left(\frac{\beta}{1+\beta} \right)^{\alpha}$}
        \KwSet{index $n$ in:}
        \hspace{1em} a. $\cC_k \in \cC$ w/pr $\frac{|\cC_k| + \alpha}{Z}$ \\
        \hspace{1.1em} b. new cluster w/pr $\frac{\alpha \overline{L}(\cX) \left(\frac{\beta}{1+\beta} \right)^{\alpha}}{Z}$
        }
        \Return{$N, \cC$}
        \caption{Sample an \gls{NSP} partition $N, \cC \sim p(N, \cC)$}
        \label{alg:sample_nsp_partition}
    \end{algorithm}
    \end{minipage}
\label{fig:sample_nsp3}
\end{figure}

\subsection{Connecting Neyman-Scott processes and Dirichlet process mixture models}
\label{sec:urn-process}

Like the~\gls{MFMM} and other nonparametric mixture models, the Neyman-Scott process can be characterized by its P\'{o}lya-urn scheme (aka ``restaurant process''). 
A P\'{o}lya-urn scheme is a discrete-time Markov process on partitions of the integers.
Indices~$n=1,2,\ldots$ are introduced one at a time, and they are either added to an existing cluster or to a new, singleton cluster. 
The Blackwell-MacQueen urn scheme~\citep{Blackwell1973} (aka ``Chinese restaurant process'') and Pitman-Yor process~\citep{Pitman1997-uy} are classic examples.
We will show that the Neyman-Scott process corresponds to an urn scheme of a very similar nature, and, like the two-parameter Pitman-Yor process, it subsumes the Blackwell-MacQueen urn scheme as a limiting case. 

The following corollary describes an urn scheme in which the marginal distribution after~$N$ steps is~$p(\cC \mid N)$.
\begin{theorem}
\label{thm:urn_process}
Let~$\cC'$ be a partition of the integers~$[N-1]$. 
The following transition operator is a distribution on partitions~$\cC$ of the integers~$[N]$ that equal~$\cC'$ when the integer~$N$ is removed.
\begin{align}
    p(\cC \mid \cC') &\propto 
    \begin{cases}
    |\cC_k| + \alpha & \text{if } \cC_k \in \cC' \text{ and } \cC_k \cup \{N\} \in \cC \\
    \alpha \frac{V_{N,|\cC'| + 1}}{V_{N,|\cC'|}} & \text{if } N \text{ is a singleton; i.e. } \{N\} \in \cC 
    \end{cases}
\end{align}
Starting with~$\cC = \{\{1\}\}$ and applying this transition operator~$N-1$ times yields a partition~$\cC$ of the integers~$[N]$ that is distributed according to~$p(\cC \mid N)$.
\end{theorem}
Theorem~\ref{thm:urn_process} is equivalent to Theorem 4.1 of~\citet{Miller2018} for mixture of finite mixture models, which is unsurprising since we have already shown that Neyman-Scott processes are special cases of \glspl{MFMM} when conditioned on the number of data points. 

Here is where the Neyman-Scott process distinguishes itself from other mixture of finite mixture models.
In the \gls{NSP}, the ratio $\sfrac{V_{N,|\cC|+1}}{V_{N,|\cC|}}$ is a constant determined only by the latent event intensity and the parameters of the weight distribution.
\begin{lemma}
\label{lem:V_ratio}
Under a Neyman-Scott process satisfying Assumption~\ref{asmp:gamma_weights}, 
\begin{align}
    \frac{V_{N,|\cC| + 1}}{V_{N,|\cC|}} &= \overline{L}(\cX) \left(\frac{\beta}{1 + \beta} \right)^{\alpha}.
\end{align}
\end{lemma}
The proof of this lemma follows from simple substitution, as shown in Appendix~\ref{app:proofs}. 
This ratio controls the probability of adding a new cluster in the urn process described in Theorem~\ref{thm:urn_process}.
In general \glspl{MFMM}, the ratio changes as a function of $|\cC|$ and $N$, and \citet{Miller2018} recommend precomputing it for a range of values.
For Neyman-Scott processes with gamma weights, it is simply a constant.

Lemma~\ref{lem:V_ratio} also hints at a close relationship between \glspl{NSP} and Dirichlet process mixture models. 
Recall that in the Blackwell-MacQueen urn scheme that generates partitions of Dirichlet process mixture model, the probability of adding a new cluster is also constant~\citep{Blackwell1973}. 
Specifically, the probability of adding to an existing cluster is proportional to the size of the cluster, $|\cC_k|$, and the probability of creating a new cluster is proportional to a constant~$\gamma$, the Dirichlet process concentration. 
The following corollary formalizes this relationship, showing how the \gls{DPMM} arises when as a limiting case of the \gls{NSP} with infinitely many latent events with weights that are almost surely zero. 

\begin{corollary}
\label{cor:nsp_dpmm}
The Neyman-Scott process with gamma weights (Assumption~\ref{asmp:gamma_weights}) approaches a Dirichlet process mixture model with concentration~$\gamma$ and a base measure with density~$p(\loc, \params)$ in the limit that~$\alpha \to 0$ while~$\alpha \, \overline{L}(\cX) \to \gamma$.
\end{corollary}

\begin{proof}
In this limit, the probability of adding a new cluster is $\lim_{\alpha \to 0} \alpha \overline{L}(\cX) (\sfrac{\beta}{1+\beta})^\alpha = \gamma$ and the probability of adding to an existing cluster is $\lim_{\alpha \to 0} |\cC_k| + \alpha = |\cC_k|$. 
Thus, the urn scheme given in Theorem~\ref{thm:urn_process} converges to the Blackwell-MacQueen urn scheme, which underlies the Dirichlet process mixture model. 
The cluster parameters in the NSP are sampled i.i.d. from $p(\loc, \params)$ in the same way cluster parameters are sampled independently from the base measure in a \gls{DPMM}.
\end{proof}

\begin{figure}[t]
    \centering
    \includegraphics[width=\linewidth]{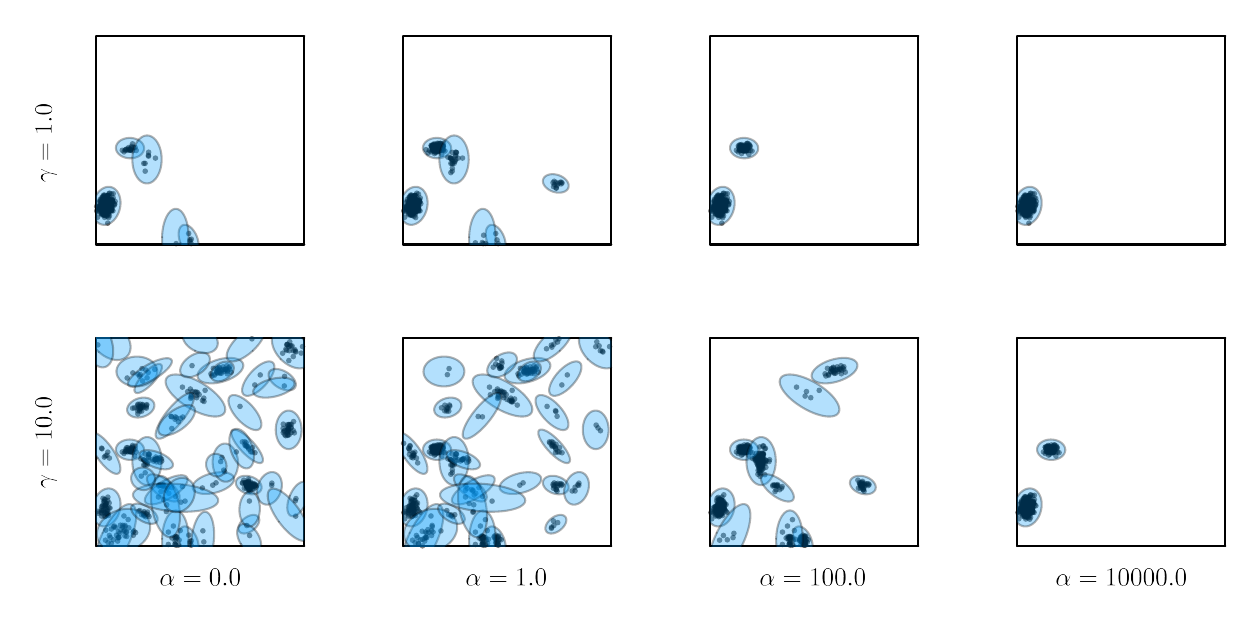}
    \caption{
    \textit{Neyman-Scott processes with gamma weights contain Dirichlet process mixture models as a limiting case.}
    Each panel shows $N=100$ data points drawn from a 2D Neyman-Scott processes with gamma weights and Gaussian clusters.
    Here, $\cX = [0,1]^2$, $\overline{L}(\cX) = \overline{\latentrate}$, $p(\loc) = \distUniform(\loc \mid \cX)$, and $\beta = 2$.
    The shape $\alpha$ of the latent event weights varies between $0$ and $10,000$, and the concentration $\gamma$ is set to $1$ (top row) or $10$ (bottom row).
    For $\alpha > 0$, the latent event intensity $\overline{\latentrate}$ is set so that $\gamma = \overline{\latentrate} \alpha \left(\sfrac{\beta}{1+\beta}\right)^\alpha$.
    When $\alpha = 0$, the data is drawn from a Dirichlet process mixture model with concentration $\gamma$.
    }
    \label{fig:nsp-data}
\end{figure}

Figure~\ref{fig:nsp-data} shows realizations of the \gls{NSP} urn scheme for~$N=100$ data points. 
In each row, the shape parameter~$\alpha$ is varied while the inverse scale~$\beta$ is fixed and the latent event rate~$\overline{\latentrate}$ is changed so that~$\gamma=\overline{\latentrate} \alpha (\sfrac{\beta}{1+\beta})^\alpha$ is held constant.
In the limit where~$\alpha \to 0$ (i.e. in the left-most column), the urn process is identical to the Blackwell-MacQueen urn process for Dirichlet process mixture models, as stated in Corollary~\ref{cor:nsp_dpmm}. 
As the shape parameter~$\alpha$ is increased, there is a greater tendency to add data points to existing clusters, leading to fewer clusters for a fixed dataset size. 
The~\gls{DPMM} limit, in contrast, has a larger number of small clusters.

The Neyman-Scott urn scheme in Theorem~\ref{thm:urn_process} is even more similar to the urn scheme of the two-parameter Pitman-Yor process~\citep{Pitman1997-uy}. 
The Pitman-Yor scheme adds indices to existing clusters with probability proportional to~$|\cC_k| - \delta$ and to a new cluster with probability proportional to~$\gamma + |\cC|\delta$.
When the \emph{discount} parameter~$\delta$ is zero, the Pitman-Yor process is also equivalent to the Dirichlet process (and its urn scheme is equivalent to the Blackwell-MacQueen urn scheme). 
When the discount parameter is negative, the Pitman-Yor urn scheme preferentially adds to existing clusters, just like in the NSP urn scheme, suggesting a correspondence between the shape parameter~$\alpha$ in the~\gls{NSP} and $-\delta$ in the Pitman-Yor scheme.
However, when~$\delta < 0$, the \emph{strength} parameter~$\gamma$ must equal~$L|\delta|$ for integer~$L$ so that the probability of creating a new cluster equals exactly zero once $|\cC|=L$.
In this case the Pitman-Yor process generates a partition for a finite Dirichlet-multinomial model. 
That is where the Neyman-Scott and Pitman-Yor urn schemes differ: in the \gls{NSP}, the probability of creating a new cluster is proportional to a constant, regardless of the number of existing clusters; in the Pitman-Yor urn scheme with a negative discount factor, that probability decreases as the number of clusters increases. 
Both the Neyman-Scott and Pitman-Yor urn schemes converge to the Blackwell-MacQueen urn scheme under certain limits, but they differ in their approach.

The connection between Neyman-Scott processes and Dirichlet process mixture models is also evident from their random measures. 
We can view the latent events of the \gls{NSP} as a discrete random measure on~$\cX \times \params$ with a Poisson-distributed number of atoms.
The $l$-th atom is located at $(\loc_l, \params_l)$ and has weight~$\weight_l$.
Intuitively, in the limit described in Corollary~\ref{cor:nsp_dpmm}, the number of atoms goes to infinity and the weights go to zero. 
In this limit, the random measure approaches a gamma process that, once normalized, yields a Dirichlet process.
The induced intensity arises by convolving this random measure with an impulse response to obtain the intensity~$\lambda(x, y \mid \{(\loc_l, \weight_l, \params_l)\}_{l=1}^L) $.
Normalizing the intensity yields the mixture density of a~\gls{DPMM}. 
We formalize this random measure perspective in Appendix~\ref{sec:random-measures}.

\section{Accounting for Background Data Points}

Often, the observed data points naturally separate into those induced by latent events and those that are ``background noise.''  
One way to account for such background events is by adding an extra term to the intensity function that produces the observed data points,
\begin{align}
    \lambda(x, y \mid \{(\loc_l, \weight_l, \params_l)\}_{l=1}^L) 
    &=
    \lambda_0(x, y) + \sum_{l=1}^L \weight_l \, p(x, y \mid \loc_l, \params_l)
\end{align}
where $\lambda_0(x, y): \cX \times \cY \to \reals_+$ is a non-negative intensity function. 
Then, the Neyman-Scott process sampling procedure (Algorithm~\ref{alg:sample_nsp1}) has one extra step of sampling background data points~$\{(x_{0,n}, y_{0,n})\}_{n=1}^{N_0} \sim \distPoissonProcess(\lambda_0(x))$. 
The complete set of data points,~$\cup_{l=0}^L  \{(x_{l,n}, y_{l,n})\}_{n=1}^{N_l}$, is the union of the background data points and those induced by latent events.

There are a few ways of modeling the background intensity. 
Without loss of generality, factor the intensity as~$\lambda_0(x, y) = \lambda_0(x) \, p(y \mid x)$, where~$p(y \mid x)$ is a normalized probability density on~$\cY$.
We will primarily focus on the spatiotemporal intensity~$\lambda_0(x)$ and assume the mark density~$p(y \mid x)$ is easy to model.
The simplest background model is a constant, homogeneous intensity,~$\lambda_0(x) = \overline{\lambda}_0$. 
Under this model, the induced data points are superimposed on top of a background of uniformly distributed data points, and the number of background data points is determined by~$\overline{\lambda}_0$.
Alternatively, $\lambda_0(x)$ could be modeled with a general, nonparametric model like a log Gaussian Cox process~\citep{moller1998log} or sigmoidal Gaussian Cox process~\citep{Adams2009-hs}.
With such models, we can incorporate any spatiotemporal covariates into the prior distribution on background intensities.
Our only requirements are that we must be able to evaluate the background intensity at any point~$(x, y)$ and the posterior distribution over background intensities given background data points be amenable to MCMC sampling. 
It is important to note that there is a risk of non-identifiability with general parametric background intensities. 
If the background intensity can subsume the impulse responses, then there is no way to distinguish between background and induced data points. 
A simple and often intuitive solution to this problem is to assume that the background intensity varies slowly relative to the impulse responses; i.e. to assume a separation of spatiotemporal scales. 

Regardless of the background model, the Neyman-Scott process with an additive background intensity and gamma weights also admits a simple partition distribution. 
Let~$\cC_0 \subseteq [N]$ denote the subset of indices assigned to the background and~$\cC$ be a partition of~$[N] \setminus \cC_0$.
\begin{theorem}
\label{thm:bkgd-partition-prior}
Under Assumption~\ref{asmp:gamma_weights}, the prior probability of the partition induced by an \gls{NSP} with background intensity~$\lambda_0(x)$, marginalizing the latent event locations, weights, and parameters, is,
\begin{align}
\label{eq:bkgd-epdist}
p(N, \cC_0, \cC) &= \left(\frac{(N-|\cC_0|)!}{N!} e^{-\weight_0(\cX)} \weight_0(\cX)^{|\cC_0|} \right) V_{N-|\cC_0|,|\cC|} \prod_{\cC_k \in \cC} \frac{\Gamma(|\cC_k| + \alpha)}{\Gamma(\alpha)},
\end{align}
where $\alpha$ and $\beta$ are the shape and rate, respectively, of the gamma prior on weights, $|\cC|$ is the number of clusters in the partition, $|\cC_k|$ is the size of the $k$-th cluster in the partition, $N = |\cC_0| + \sum_{\cC_k \in \cC} |\cC_k|$ is the total number of data points (a random variable), and
where $\weight_0(\cX) = \int_{\cX} \lambda_0(x) \dif x$ is the integrated background intensity. 

A partition of size~$N$ can be generated recursively with the following urn scheme,
\begin{align}
    p(\cC_0, \cC \mid \cC_0', \cC') &\propto 
    \begin{cases}
    \weight_0(\cX) (1 + \beta) & \text{if } \cC_0 = \cC_0' \cup \{N\} \text{ and } \cC = \cC' \\
    |\cC'_k| + \alpha & \text{if } \cC'_k \in \cC' \text{ and } \cC'_k \cup \{N\} \in \cC\\
    \alpha \overline{L}(\cX) \left(\frac{\beta}{1+\beta} \right)^\alpha & \text{if } N \text{ is a singleton; i.e. } \{N\} \in \cC
    \end{cases}
\end{align}
where~$\{\cC_0'\} \cup \cC'$ is a partition of the integers~$[N-1]$ and~$\{\cC_0\} \cup \cC$ is a partition of~$[N]$ obtained by adding index~$N$ as described above. 
\end{theorem}
Theorem~\ref{thm:bkgd-partition-prior} closely parallels Theorems~\ref{thm:partition-prior} and~\ref{thm:urn_process}, as does its proof (given in Appendix~\ref{app:proofs}). 
The only difference is that here we separate the background indices,~$\cC_0$, from the partition of the remaining indices,~$\cC$.
In the terminology of~\citet{pitman2006},  eq.~\eqref{eq:bkgd-epdist}, once normalized by $p(N)$, is a \textit{partially} exchangeable partition probability function since it is not symmetric in the cluster indices; the background cluster is treated differently.
\section{Bayesian Learning and Inference for Neyman-Scott Processes}
\label{sec:inference}

Having shown the relationship between Neyman-Scott processes with gamma weights, mixture of finite mixture models, and Dirichlet process mixture models, we can adapt standard collapsed Gibbs sampling algorithms for \glspl{MFMM}~\citep{Miller2018} and \glspl{DPMM}~\citep{Maceachern1994-fy, Neal2000} to \glspl{NSP}. 
The resulting algorithm is quite different from existing inference algorithms for \glspl{NSP}. 
Rather than proposing to add or remove latent events, as in reversible jump MCMC algorithms~\citep{Moller2003,Moller2014}, the collapsed Gibbs algorithm marginalizes over latent event locations and parameters and operates directly on the posterior over partitions. 
Unlike minimum contrast estimation methods, the collapsed Gibbs algorithm asymptotically generates samples from the posterior distribution and makes minimal assumptions about the latent events and their impulse responses. 


\begin{algorithm}[t]
\KwIn{Data points $\{(x_n, y_n)\}_{n=1}^N$, hyperparameters~$\alpha, \beta, \overline{L}(\cX)$}
Initialize all data points to the background: $\cC_0 = \{1,\ldots,N\}$ and $\cC = \{\ \}$. \\
\SetKwBlock{Loop}{repeat}{end}
\Loop($S$ times to draw $S$ samples){
  1. Sample parent assignments, integrating over latent events:
  
  \hspace{1.5em} \textbf{for} $n = 1,\hdots , N$, remove $n$ from its current cluster and place it in...\\
  \hspace{2.5em} a. the background cluster, $\cC_0$, with probability $\propto \lambda_0(x_n, y_n) (1+\beta)$ \\
  \hspace{2.5em} b. cluster $\cC_k$, with probability $\propto (|\cC_k| + \alpha) \, p(x_n, y_n \mid \{(x_{n'}, y_{n'}) : n' \in \cC_k\})$ \\
  \hspace{2.5em} c. a new cluster with probability $\propto \alpha \, \overline{L}(\cX) \left(\frac{\beta}{1+\beta} \right)^\alpha p(x_n, y_n)$ \\
  \hspace{1.5em} \textbf{end}
 
  \vspace{.6em}
  2. Sample background intensity given the data points $\{(x_n, y_n): n \in \cC_0\}$.
  
  \vspace{.6em}

  3. Sample latent events: \\
  \hspace{1.5em} \textbf{for} $\cC_k \in \cC$: \\
  \hspace{2.5em} Sample locations and parameters $\loc_k, \params_k \sim p(\loc_k, \params_k \mid \{(x_n, y_n): n \in \cC_k\})$ \\
  \hspace{2.5em} Sample weights $\weight_k \sim \distGamma(\alpha + |\cC_k|, \beta + 1)$ \\
  \hspace{1.5em} \textbf{end}
  
  \vspace{.6em}

  4. [Optionally] Sample hyperparameters: \\
  \hspace{1.5em} Sample hyperparamters of~$p(\loc, \params)$ given the samples~$\{(\loc_k, \params_k)\}$ \\
  \hspace{1.5em} Sample number of empty clusters $E \sim \distPoisson \left(\overline{L}(\cX) \left(\frac{\beta}{1+\beta}\right)^\alpha \right)$. Set $L = |\cC| + E$. \\
  \hspace{1.5em} Sample homogenous latent event intensity $\overline{\nu} \sim \distGamma(\alpha_\latentrate + L, \beta_\latentrate + |\cX|)$ \\
  \hspace{1.5em} Sample weights for empty clusters $\weight_l \sim \distGamma(\alpha, \beta + 1)$ for $l=|\cC| + 1,\ldots,L$\\
  \hspace{1.5em} Sample hyperparamters $p(\alpha, \beta \mid \{\weight_l\}_{l=1}^{L}) \propto p(\alpha, \beta) \prod_{l=1}^L \distGamma(\weight_l \mid \alpha, \beta)$\\

  \vspace{.6em}

  5. [Optionally] Update partition with split-merge moves (see~\Cref{app:split-merge}) \\

}
\caption{Collapsed Gibbs sampling for Neyman-Scott processes with gamma weights}
\label{alg:collapsedgibbs}
\end{algorithm}

Algorithm~\ref{alg:collapsedgibbs} assumes access to the marginal likelihood,
\begin{align}
    \label{eq:marginal}
    p(x_n, y_n) &= \int p(x_n, y_n \mid \loc, \params) \, p(\loc, \params) \dif \loc \dif \params
\end{align}
and the predictive likelihood
\begin{align*}
    p(x_n, y_n \mid \{(x_{n'}, y_{n'}) : n' \in \cC_k\}) &=
    \int p(x_n, y_n \mid \loc, \params) \, p(\loc, \params \mid \{(x_{n'}, y_{n'}) : n' \in \cC_k\}) \dif \loc \dif \params.
\end{align*}
Closed-form expressions for these likelihoods are available for conjugate exponential family models.
However, extensions for collapsed Gibbs sampling in nonconjugate \glspl{DPMM} and \glspl{MFMM} also apply here.
If the marginal likelihood in \cref{eq:marginal} is not available in closed form, auxiliary variable methods like Algorithm 8 of \citet{Neal2000} may be used.  
Likewise, split-merge methods~\citep{Jain2004, Jain2007} can improve the mixing time of the sampling algorithm with larger updates to the partition. 
Further details on the sampling steps, hyperparameter selection, initialization, and parallelization are in Appendix~\ref{app:inference}.

\glsreset{NSP}
\glsreset{DPMM}

\section{Experiments}

\begin{figure}[t]
\centering
\includegraphics[width=\linewidth]{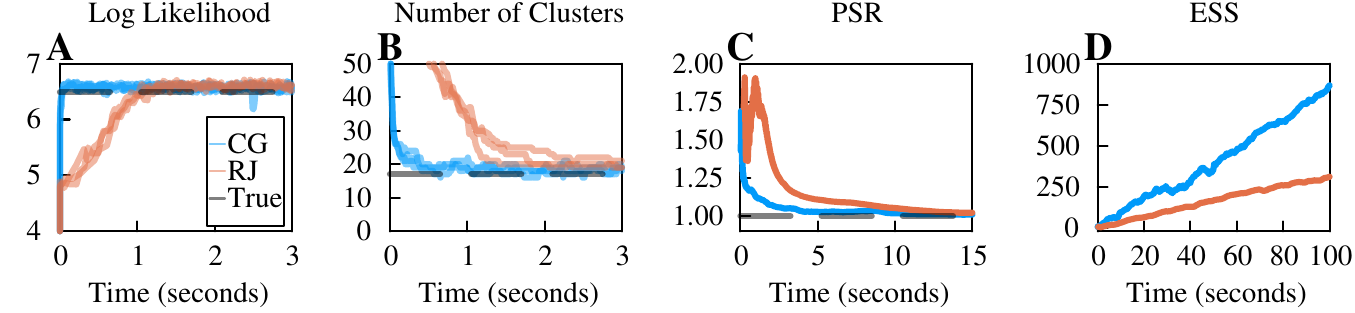}
\caption{
\textit{Performance of the collapsed Gibbs (CG) and reversible jump MCMC (RJ) samplers.}
\textbf{(A-B)} Sampler trace plots of log-likelihood and inferred number of non-empty clusters.
\textbf{(C)} Potential scale reduction factor (PSR), also known as $\hat R$, for the number of non-empty clusters.
A PSR close to $1.0$ indicates convergence.
\textbf{(D)}
Effective sample size (ESS) for the number of non-empty clusters.
ESS is a metric of sampler efficiency.
}
\label{fig:cg-rj}
\end{figure}

We study the collapsed Gibbs algorithms for the Neyman-Scott process model on synthetic data as well as real neural recordings and document streams.
All of the code for our experiments is written in the Julia programming language and made publicly available in the Github repository \texttt{redacted.jl}.
Across datasets, we find that the \gls{NSP} offers a compelling alternative to existing clustering models in tasks like sequence detection in neural spike train data and event detection in document streams.

\subsection{Sampler convergence and performance}
\label{sec:sampler-experiment}

In our first experiment, we compare the mixing time and sampler efficiency of our collapsed Gibbs sampler with split-merge moves to a standard reversible jump MCMC sampler on a synthetic dataset.
Overall, we find that the collapsed Gibbs sampler converges to the posterior more quickly and produces more effective samples per second than the reversible jump sampler.

We generate synthetic data from a 2D Neyman-Scott process across the unit square $\cX = [0, 1]^2$ with gamma weights and Gaussian clusters. Specifically, we set
\begin{align}
    \nu(\loc, \weight, \params) &= \overline{\nu} \, \distGamma(\weight \mid \alpha, \beta) \, \distInvWishart(\params \mid d, \Psi) \\
    \lambda (x \mid \{\loc_l, w_l, \params_l\}_{l=1}^L )
    &= \sum_{l=1}^L w_l \, \distNormal(x \mid \loc_l, \params_l)
\end{align}
where~$\overline{\nu}$ specifies the rate of latent events and $\distInvWishart(\cdot \mid d, \Psi)$
denotes an inverse Wishart density with~$d$ degrees of freedom and scale matrix~$\Psi$.
We set the background rate to zero for simplicity.
\Cref{fig:nsp-dpm}A shows a sample from this~\gls{NSP}.

We then fit an \gls{NSP} to the generated data using the same priors.
We initialize each Markov chain by sampling each of the $N$ parent assignments uniformly at random from the set $\{1, \ldots, N\}$ (see Appendix~\ref{app:inference} for further detail).
The reversible jump MCMC sampler alternates between proposing latent event birth-death moves, sampling parent assignments conditioned on the latent event parameters, and sampling of the latent event parameters conditioned on parent assignments.
Analogously, the collapsed Gibbs sampler alternates between sampling parent assignments and proposing split-merge moves, both marginalized over latent event parameters.
The split-merge moves propose either splitting a cluster in two or merging an existing cluster, operating directly on the posterior over partitions.
These updates, like the collapsed Gibbs updates, come directly from our derivation of the posterior probability over partitions
(see Appendix~\ref{app:inference} for further detail).

We run 3 independent chains of each sampler for 2 minutes ($\approx 7{\small,}000$ samples each) and plot their performance in \Cref{fig:cg-rj}.
In Panels~A~and~B, we display trace plots of joint log-likelihood and and the inferred number of non-empty clusters during inference.
In Panel~C, we plot the potential scale reduction factor (PSR, also known as $\hat R$) of the number of non-empty clusters.
Both samplers converge to PSR near 1.0, indicating that the independent chains have converged to the same posterior mode.
Finally, Panel~D shows the effective sample size for each sampler.
The collapsed Gibbs sampler mixes significantly more rapidly than the reversible jump MCMC sampler.

\subsection{Comparison with DPMM}

Our second experiment compares the performance of the \gls{NSP} with the \gls{DPMM} on the same synthetic dataset.
Our results indicate that a \gls{DPMM} cannot model data from an \gls{NSP} well: it leads to inaccurate cluster assignments and a biased estimate of the number of clusters.

\begin{figure}[t]
\centering
\includegraphics[width=\linewidth]{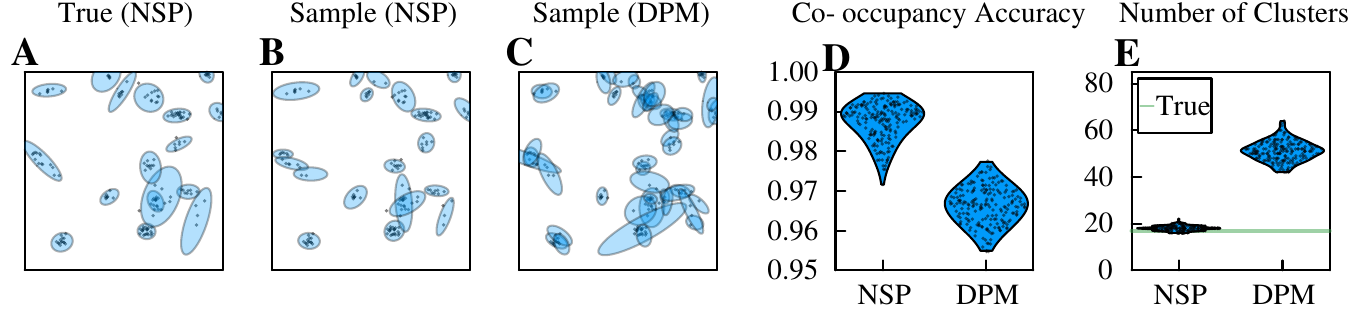}
\caption{
\textit{Performance of the \gls{NSP} and the \gls{DPMM} on synthetic 2D \gls{NSP} data.}
\textbf{(A)} True data generated by a Neyman-Scott process.
\textbf{(B)} Inferred parent assignments from a single sample of the posterior of the Neyman-Scott process, as estimated using the proposed MCMC algorithm.
\textbf{(C)} Same as (B), but under the \gls{DPMM} posterior.
\textbf{(D-E)} Distribution latent event assignment co-occupancy accuracy (D) and inferred number of clusters (E) for the NSP and DPM.
Data from the last 100 samples of all three chains is displayed.
As expected, the DPM significantly overestimates the true number of clusters.
}
\label{fig:nsp-dpm}
\end{figure}

We use the same synthetic data and collapsed Gibbs sampler as in Section~\ref{sec:sampler-experiment} to fit both models, running 3 independent chains per model.
Figure~\ref{fig:nsp-dpm}, panels~B~and~C show the resulting parent assignments at the end of sampling one of the three chains, for each the NSP and the \gls{DPMM}.
We observe that the inferred assignments using the \gls{NSP} qualitatively outperform those of the \gls{DPMM}.
To quantitatively assess the accuracy of the inferred parent assignments, we use \textit{co-occupancy accuracy}.
Let~$z$ and~$z'$ denote the parent assignments obtained from partitions~$\cC$ and~$\cC'$, assuming arbitrary ordering of their clusters.
The co-occupancy accuracy is defined as,
    $\mathbf{accuracy}(z, z')
    = \frac{1}{N^2}
    \sum_{n=1}^N \sum_{m=1}^N
    \bbI[z_n = z_m] \bbI[z'_n = z'_m]
    + \bbI[z_n \neq z_m] \bbI[z_n' \neq z_m'],$
where~$\bbI[\cdot]$ is the binary indicator function.
Panel~D shows the co-occupancy accuracy of the resulting partition for the last 100 samples of each chain.
Finally, Panel~E shows the inferred number of clusters over the same duration.
The \gls{NSP} infers both co-occupancy and the number of clusters more accurately than the \gls{DPMM}.
These results suggest that the \gls{DPMM} is not an appropriate model for fitting data generated by an \gls{NSP} (or an \gls{MFMM}, see \cite{Miller2018} for additional examples).

\subsection{Application: Detecting sequences in neural spike trains}
\label{sec:spike_trains}
We next apply the Neyman-Scott process model to perform sequence detection in neural data.
With multielectrode arrays, neuroscientists can record the times at which individual neurons fire action potentials, or \textit{spikes}.
A collection of spike times from one or more neurons is called a \textit{spike train}.
In many experimental settings, groups of neurons (sometimes called ``ensembles'') are hypothesized to fire repeatedly in sequences~\citep[e.g.]{Hahnloser2002}.
Identifying these sequences in spike train data is a longstanding problem that has spurred a wide variety of modeling approaches \citep{Abeles2001,Russo2017,Quaglio2018,Mackevicius2019}.
The results below extend~\citet{Williams2020_ppseq}.

We model the spike train as a marked point process generated by the Neyman-Scott process described in Section~\ref{sec:nsp}.
Each sequence corresponds to a latent event, which consists of a time~$\loc_l \in [0,T]$, weight~$\weight_l \in \reals_+$, and a discrete type~$\params_l \in\{1, \ldots, S\}$.
We model the latent event intensity as, ${\nu(\loc, \weight, \params) = \overline{\nu} \, \distGamma(\weight \mid \alpha, \beta) \, \distCategorical(\params \mid \pi)}$,
where~$\pi \in \Delta_S$ is a distribution over sequence types.
We model the conditional intensity of the observed spikes as,
\begin{align}
\lambda(x, y \mid \{\loc_l, w_l, \params_l\}_{l=1}^L )
&= \overline{\lambda}_0 \, \distCategorical(y \mid a_0) + \sum_{l=1}^L w_l \, \distCategorical(y \mid a_{\params_l}) \, \cN(x \mid \loc_l + b_{y, \params_l}, \sigma_{y,  \params_l}^2),
\end{align}
where~$\overline{\lambda}_0$ sets the rate of background spikes across all neurons,~$a_0 \in \Delta_Y$ is a distribution over neurons,~${a_s \in \Delta_Y}$ is a distribution over neurons for spikes in sequences of type~$s$, and ${b_{y,s} \in \mathbb{R}}$ and~${\sigma_{y,s} \in \reals_+}$ specify the latency and width, respectively, of impulse responses induced on neuron~$y$ by latent events of type~$s$.



\begin{figure}[t]
\centering
\includegraphics[width=\linewidth]{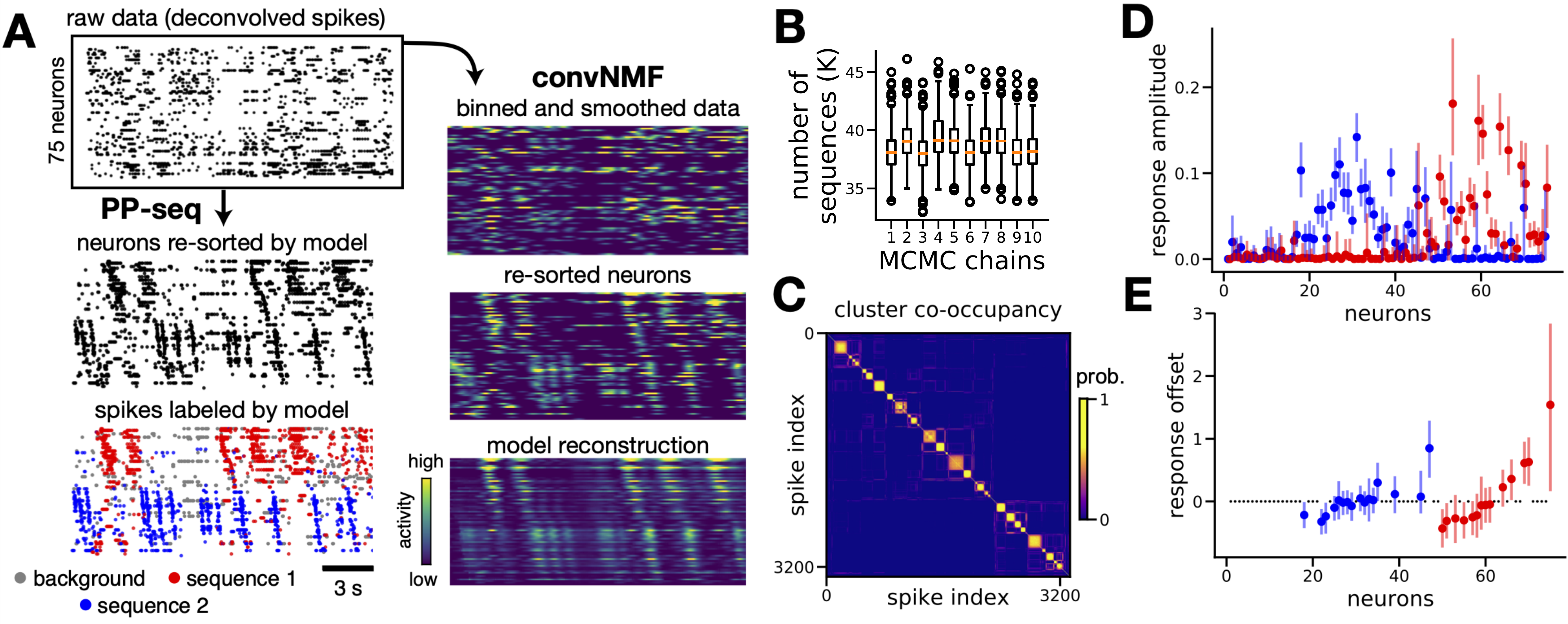}
\caption{
\textit{Zebra Finch HVC data.}
\textit{(A)} Raw spike train (top) and sequences revealed by the Neyman-Scott process (left) and convNMF (right).
\textit{(B)} Box plots summarizing samples from the posterior on number of sequences, $K$, derived from 10 independent MCMC chains.
\textit{(C)} Co-occupancy matrix summarizing probabilities of spike pairs belonging to the same sequence.
\textit{(D)} Credible intervals for evoked amplitudes for sequence type 1 (red) and 2 (blue).
\textit{(E)} Credible intervals for response offsets (same order and coloring as \textit{D}). Estimates are suppressed for small-amplitude responses (gray dots).
}
\label{fig:bird_fig}
\end{figure}

\label{subsec:bird}

We applied the Neyman-Scott process to a recording of higher vocal center (HVC) premotor neurons in a zebra finch,
which generate sequences that are time-locked to syllables in the bird's courtship song.
\Cref{fig:bird_fig}A shows a qualitative comparison of the performance of a discrete time method called convNMF~\citep{Mackevicius2019} and the Neyman-Scott process for neural sequence detection.
The raw data (top panel) showed no visible spike patterns; however, clear sequences were revealed by sorting the neurons lexographically by preferred sequence type and the temporal offset parameter, as inferred by the Neyman-Scott process.
While both models extracted similar sequences, the Neyman-Scott process provided a finer scale annotation of the final result, offering, for example, attributions at the level of individual spikes to sequences.

Further, the Neyman-Scott process offered uncertainty estimates for key parameters via the MCMC samples.
\Cref{fig:bird_fig}B summarizes uncertainty in the total number of latent events pooled over 10 independent MCMC runs with different random seeds---the chains differ slightly in their estimates, with medians differing by 1-2 sequences across chains.
The uncertainty is largely due to the rapid sequences (in blue) shown in panel A.
\Cref{fig:bird_fig}C displays a co-occupancy matrix where element~$(i,j)$ corresponds to the probability that spike~$i$ and spike~$j$ were attributed to the same sequence.
Finally, \cref{fig:bird_fig}D-E shows the amplitude and offset for each neuron's sequence-evoked response with 95\% posterior credible intervals.
These results naturally follow from the probabilistic construction of the the Neyman-Scott process model.


\subsection{Application: Detecting world events in streams of diplomatic cables}

\label{subsec:document-model}

Detecting important events from historical documents is a common task for historians \citep{howell2001reliable}, and large-scale digital corpora offer new challenges and opportunities in this domain.
Here we develop a Neyman-Scott process model to perform event detection in document streams.
We study a dataset of US State Department declassified cables from June 21--July 31, 1976~\citep{connelly2021diplomatic}, which spans a duration of $T=40$ days and totals $N=34,732$ cables between~$A=2,360$ entities.
Each cable has a time stamp~$x_n \in [0, T]$ and a mark~$y_n = (y_n^{(a)}, y_n^{(c)})$, where~$y_n^{(a)} \in [A]$ specifies the entity and $y_n^{(c)} \in \naturals_0^V$ is a vector of word counts for vocabulary of size~$V=21,819$.

To identify historical events with the cables dataset, we posit that: (1) each entity sends out \textit{background cables} about regional affairs, e.g.\ a Thailand entity regularly sends out cables about affairs in Southeast Asia; and (2) when a salient event occurs, some entities will send out \textit{event cables}. With this intuition, we model the cables with the Neyman-Scott process described in Section~\ref{sec:examples}.
The observed cables are driven by latent ``world events'' with times~$\loc_l \in [0,T]$, weights~$\weight_l \in \reals_+$, and parameters~$\params_l = (\params_l^{(a)}, \params_l^{(c)}) \in \Delta_A \times \reals_+^V$ that determine the distribution over authors and words.
The latent event rate is modeled as,
\begin{align}
    \nu(\loc, \weight, \params)
    &= \overline{\latentrate} \, \distGamma(\weight \mid \alpha, \beta) \, \distDirichlet(\params^{(a)} \mid \alpha_a) \, \prod_{v=1}^V \distGamma(\params^{(c)}_{v} \mid \alpha_c, \beta_c).
\end{align}
The only new parameters are~$\alpha_a$, the concentration parameter of the author distribution, $\alpha_c$ and~$\beta_c$, the concentration and rate parameters of the word intensity distribution.

The conditional intensity of the observed events is modeled as,
\begin{multline}
    \lambda(x, y \mid \{\loc_l, \weight_l, \params_l\}_{l=1}^L )
    = \overline{\lambda}_0 \, \distCategorical(y^{(a)} \mid \theta_0^{(a)}) \, \prod_{v=1}^V \distPoisson(y_v^{(c)} \mid \phi_{y^{(a)},v}) \\
    + \sum_{l=1}^L w_l \, \cN(x \mid \loc_l, \sigma^2) \, \distCategorical(y^{(a)} \mid \params_l^{(a)}) \, \prod_{v=1}^V \distPoisson(y^{(c)}_{v} \mid \params_{l,v}^{(c)}).
\end{multline}
where~$\overline{\lambda}_0$ sets the rate of background cables,~$\theta_0^{(a)} \in \Delta_A$ is a distribution over background cable authors, and~$\phi_{a,v}$ specifies the rate of author~$a$ using word~$v$ in its background cables.
Our goal is to infer putative world events, as represented by the timestamps~$m_l$ and parameters~$\theta_l$.


We compared this Neyman-Scott process model for document streams to a baseline clustering model on this dataset similar to the model proposed by~\citet{Chaney:2016}.
Both models allow for background cables with sender-specific rates and word distributions.
The models differ in how they handle latent world events.
Whereas the Neyman-Scott process models the cables data as a continuous-time process with latent events, the baseline model bins the time into one-week intervals and allows one latent event per week.
Thus, the baseline model must necessarily combine contemporaneous world events into one cluster.

To compare the Neyman-Scott process model and the baseline model quantitatively, we compare the predictive log-likelihoods produced on held-out regions of the data.
In particular, we hold out 10\% of the event space by masking 10\% of the observed time interval (4 randomly chosen days) for each embassy during inference.
The baseline model achieves a predictive likelihood of $-196.9$ nats/document whereas the \gls{NSP} achieves -$189.1$ nats/document, suggesting the \gls{NSP} better generalizes to out-of-sample data.

\begin{figure}
\centering
\includegraphics[width=5.5in,]{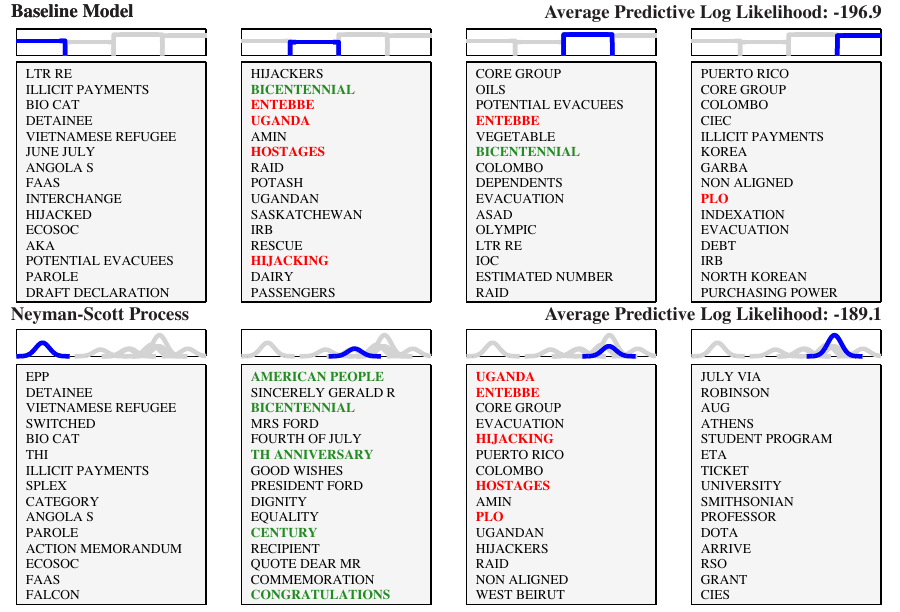}
\caption{
\textit{Document stream detection model.}
The baseline model with one cluster per week (top) necessarily cables overlapping world events, like the American Bicentennial and Operation Entebbe, both of which occurred on July 4, 1976. By contrast, Neyman-Scott process (bottom) successfully distentangles these contemporaneous events. For more information on this dataset, see \texttt{http://www.history-lab.org}.
}
\label{fig:embassy}
\end{figure}

\Cref{fig:embassy} presents the important diplomatic events detected by both models.
The Neyman-Scott process model inferred two important and contemporaneous events: one about a celebration in the United States (green) and another about a hostage event involving Uganda (red), as seen in the top words shown in~\cref{fig:embassy} (bottom).
The first event points to United States Bicentennial, which celebrated the 200th anniversary of adopting the Declaration of Independence~\citep{halloran_1976}. The second event points to Operation Entebbe, which was a hostage-rescue mission in Uganda ordered by Israel Defence Forces~\citep{smith_1976}.

Notably, the Neyman-Scott process model was able to distinguish temporally overlapping events.
It inferred that the two important events detected were temporally overlapping.
This inference was consistent with historical facts: both the Bicentennial celebration and Operation Entebbe indeed occurred on the same day in history (July 4, 1976).
In contrast, the baseline clustering model (\Cref{fig:embassy}, top) merged the Bicentennial celebration event and Operation Entebbe since they both happened in the same week
This result illustrates the modeling power of Neyman-Scott processes for document streams: its continuous-time nature allows for the detection of overlapping latent events with temporal, entity, and content specificity.


\section{Discussion}
\label{sec:conclusion}

\glsreset{NSP}
\glsreset{DPMM}
\glsreset{MFMM}

\Glspl{NSP} are doubly stochastic point processes that generated clusters of data points.
Like \glspl{DPMM} and \glspl{MFMM}, the number of clusters is a random variable. 
\glspl{NSP}, however, probabilistically model the number of observed  within an observed spatiotemporal region, making them well suited to many applications in spatial statistics.

Traditionally, parameter estimation and latent event inference in Neyman-Scott processes have been done via moment-matching or reversible jump MCMC methods. 
The former is limited in scope, requiring strong assumptions about the nature of the latent events and their impulse responses; the latter is quite general, but its performance is limited by our ability to craft efficient Metropolis-Hastings proposals.
A separate line of work has developed computational methods for multiplicative intensity models~\citep{ishwaran2004computational} and a general framework for models derived from Poisson random measures~\citep{james2005bayesian}, which are important precursors to our work.
Here, we proposed a novel collapsed Gibbs sampling algorithm for Neyman-Scott processes with gamma weights (Assumption~\ref{asmp:gamma_weights}), an assumption that still permits a wide family of models, as showcased in our experiments. We also showed how Neyman-Scott processes with gamma weights are intimately related to mixture of finite mixture models and Bayesian nonparametric mixture models derived from the Dirichlet process and the Pitman-Yor process.



Our experiments highlighted the breadth of spatiotemporal clustering problems amenable to Neyman-Scott process models. 
In addition to contrasting \glspl{NSP} and \glspl{DPMM} with synthetic data and demonstrating the latter's tendency to over-segmentation, we also showed how \glspl{NSP} are useful for neural sequence detection and event detection in document streams.
These are only two example applications, and it is easy to envision applications in other domains like ecology (e.g. modeling plant locations or rainfall distributions), epidemiology (e.g. modeling disease outbreaks), astronomy (e.g. modeling galaxy locations in survey data), seismology (e.g. modeling earthquake occurrences), and more.

Future work could aim to relax the key assumption of this paper---namely, that the latent events contain independent, gamma-distributed weights---with a model that allows for some dependency. 
Recent work on repulsive point processes may offer some ways of addressing this limitation~\citep{ghilotti2023bayesian}.
Likewise, recent work on deep \glspl{NSP} and variational inference for \glspl{NSP} offers more expressive models and a greater variety of inference algorithms~\citep{hong2021deep,hong2023variational}. 
With better algorithms for posterior inference in Neyman-Scott processes and clearer understanding of their relationship to Bayesian nonparametric mixture models, many generalizations and applications may follow.

\singlespacing

\begin{small}
  \subsection*{Acknowledgements}
  We thank David Blei, Peter Orbanz, and Lancelot James for helpful discussions and suggestions. 
  We also thank Allison Chaney and Matthew Connelly for assistance with the US State Department cables dataset.
  YW is supported by the Office of Naval Research under grant number N00014-23-1-2590 and the National Science Foundation under Grant number 2231174 and number 2310831
  AD is supported by the U.S. Department of Energy, Office of Science, Office of Advanced Scientific Computing Research, Department of Energy Computational Science Graduate Fellowship under Award Number DE-SC0021110.
  SWL is supported by grants from the NIH Brain Initiative (U19NS113201, R01NS113119, R01NS130789, and R01NS131987), the National Science Foundation (2223827), the Simons Foundation (SCGB 697092 and 294338), the Sloan Foundation, and the McKnight Foundation.
\end{small}

\newpage
\appendix

\doublespacing

\section{Proofs}
\label{app:proofs}
\begin{proof}[Proof of Theorem~\ref{thm:partition-prior}]
Our derivation follows the approach of~\citet{Miller2018} for mixture of finite mixture models, but we adapt it to Neyman-Scott processes with gamma weights.
The major difference is that the number of data points is a random variable in the \gls{NSP}, so the distribution above is over partitions of random size.
First, return to the generative process described in Algorithm~\ref{alg:sample_nsp3} and integrate over the latent event amplitudes to obtain the marginal distribution on latent cluster sizes given the total number of latent events~$L$.  We have,
\begin{align*}
    p(N_1, \ldots, N_L \mid L)
    &= \prod_{l=1}^L \int \distPoisson(N_l \mid \gamma_l) \, \mathrm{Ga}(\gamma_l \mid \alpha, \beta) \dif \gamma_l \\
    &= \prod_{l=1}^L \distNegBinomial(N_l \mid \alpha, (1+\beta)^{-1}) \\
    &= \prod_{l=1}^L \frac{\Gamma(N_l + \alpha)}{N_l! \, \Gamma(\alpha)} \left( \frac{\beta}{1+\beta} \right)^\alpha \left(\frac{1}{1+\beta} \right)^{N_l} \\
    &= \left( \frac{\beta}{1+\beta} \right)^{L\alpha} \left( \frac{1}{1+\beta} \right)^{N} \prod_{l=1}^L \frac{\Gamma(N_l + \alpha)}{N_l! \, \Gamma(\alpha)} .
\end{align*}
Let~$z_n \in \{1,\ldots,L\}$ denote the parent assignment of the~$n$-th data point. There are~$\binom{N}{N_1,\ldots,N_L}$ parent assignments consistent with the latent cluster sizes~$N_1,\ldots,N_L$, and they are all equally likely under the prior, so the conditional probability of the parent assignments is,
\begin{align*}
    p(\{z_n\}_{n=1}^N \mid L) 
    &= \binom{N}{N_1, \ldots, N_l}^{-1} \, \left( \frac{\beta}{1+\beta} \right)^{L\alpha} \left( \frac{1}{1+\beta} \right)^{N} \prod_{l=1}^L \frac{\Gamma(N_l + \alpha)}{N_l! \, \Gamma(\alpha)} \\
    &= \frac{1}{N!} \left( \frac{\beta}{1+\beta} \right)^{L\alpha} \left( \frac{1}{1+\beta} \right)^{N} \prod_{l=1}^L \frac{\Gamma(N_l + \alpha)}{\Gamma(\alpha)}
\end{align*}

The parent assignments above induce a partition, but technically they assume an ordering of the latent events.  Moreover, if some of the latent events fail to produce any observed events, they will not be included in the partition.  In performing a change of variables from parent assignments~$\{z_n\}_{n=1}^N$ to partitions~$\cC$, we need to sum over latent event assignments that produce the same partition.  There are~${\binom{L}{|\cC|} |\cC|! = \frac{L!}{(L-|\cC|)!}}$ such assignments if there are~$L$ latent events but only~$|\cC|$ parts in the partitions. Thus,
\begin{align}
\label{eq:p_NC_given_L}
p(N, \cC \mid L)
&= \frac{L!}{(L-|\cC|)!} \frac{1}{N!} \left( \frac{\beta}{1+\beta} \right)^{L\alpha} \left( \frac{1}{1+\beta} \right)^{N} \prod_{\cC_k \in \cC} \frac{\Gamma(|\cC_k| + \alpha)}{\Gamma(\alpha)}.
\end{align}
Clearly,~$L$ must be at least~$|\cC|$ in order to produce the partition. 

Finally, we sum over the number of latent events~$L$ to obtain the marginal probability of the partition,
\begin{align*}
p(N, \cC) &= \sum_{L=|\cC|}^\infty \distPoisson(L \mid \overline{L}(\cX)) \, p(N, \cC \mid L) \\
&= V_{N,|C|} \prod_{\cC_k \in \cC} \frac{\Gamma(|\cC_k|+\alpha)}{\Gamma(\alpha)},
\end{align*}
where 
\begin{align}
\label{eq:V}
V_{N,|\cC|} &= \frac{1}{N!} \left(\frac{1}{1+\beta} \right)^{N} \sum_{L=|\cC|}^\infty \distPoisson(L \mid \overline{L}(\cX)) \, \frac{L!}{(L-|\cC|)!} \left(\frac{\beta}{1+\beta}\right)^{L \alpha}.
\end{align}
\end{proof}

\begin{proof}[Proof of Theorem~\ref{thm:urn_process}]
Let~$\cC$ be a partition of~$\{1,\ldots,N\}$ and~$\cC \setminus N$ denote the partition of~$\{1,\ldots,N-1\}$ obtained after removing~$N$ from~$\cC$.
The transition operator defined in the theorem guarantees that~$\cC' = \cC \setminus N$.
We will proceed by induction. 
Suppose~$\cC'$ is distributed as~$p(\cC' \mid N-1)$.
The marginal distribution of the partition~$\cC$ obtained by the transition operator is,
\begin{align}
    q(\cC) &= \sum_{\cC'} p(\cC' \mid N-1) \, p(\cC \mid \cC') \\
    &= \frac{V_{N-1,|\cC \setminus N|}}{p(N-1)} \prod_{\cC_k \in \cC \setminus N} \left[\frac{\Gamma(|\cC_k| + \alpha)}{\Gamma(\alpha)} \right] p(\cC \mid \cC \setminus N).
\end{align}
There are two cases to consider. 
First, suppose~$N$ was added to an existing cluster (i.e.~$\cC_k \in \cC \setminus N$ and $\cC_k \cup \{N\} \in \cC$). 
Then,
\begin{align}
    q(\cC) &\propto \frac{V_{N-1,|\cC \setminus N|}}{p(N-1)} \prod_{\cC_k \in \cC \setminus N} \left[ \frac{\Gamma(|\cC_k| + \alpha)}{\Gamma(\alpha)} \right] (|\cC_k| + \alpha) \\
    &\propto \frac{V_{N-1,|\cC \setminus N|}}{p(N-1)} \prod_{\cC_k \in \cC} \frac{\Gamma(|\cC_k| + \alpha)}{\Gamma(\alpha)}
    \label{eq:case1}
\end{align}
Second, suppose that a new cluster was created (i.e.~$\{N\} \in \cC$). 
Then,
\begin{align}
    q(\cC) &\propto \frac{V_{N-1,|\cC \setminus N|}}{p(N-1)} \prod_{\cC_k \in \cC \setminus N} \left[ \frac{\Gamma(|\cC_k| + \alpha)}{\Gamma(\alpha)} \right] \left(\alpha \frac{V_{N,|\cC \setminus N| + 1}}{V_{N,|\cC \setminus N|}}\right) \\
    &\propto \frac{V_{N-1,|\cC \setminus N|}}{p(N-1)} \frac{V_{N,|\cC|}}{V_{N,|\cC \setminus N|}} \prod_{\cC_k \in \cC} \frac{\Gamma(|\cC_k| + \alpha)}{\Gamma(\alpha)}
    \label{eq:case2}
\end{align}
Multiplying equations~\eqref{eq:case1} and~\eqref{eq:case2} by~$\frac{V_{N,|\cC \setminus N|} p(N-1)}{V_{N-1,|\cC \setminus N|}}$ and noting that~$|\cC \setminus N| = |\cC|$ when~$N$ is added to an existing cluster, the probability in both cases reduces to
\begin{align}
    q(\cC) &\propto V_{N,|\cC|} \prod_{\cC_k \in \cC} \frac{\Gamma(|\cC_k| + \alpha)}{\Gamma(\alpha)}
\end{align}
in both cases.
Since this distribution is proportional to eq.~\eqref{eq:epdist} and~$\cC$ is a partition of~$\{1,\ldots,N\}$, the normalizing constant must be~$p(N)$.
Thus,~$q(\cC) = p(\cC \mid N)$.

To complete the proof, let's double check that the base case~$\cC = \{\{1\}\}$ is indeed distributed as~${p(\cC \mid N=1)}$,
\begin{align}
    p(\cC = \{\{1\}\} \mid N=1) &= \frac{V_{1,1}}{p(N=1)} \cdot \frac{\Gamma(1 + \alpha)}{\Gamma(\alpha)} \\
    &= \frac{(1+\beta)^{-1} \sum_{L=1}^\infty \distPoisson(L \mid \overline{L}(\cX)) \, \frac{L!}{(L-1)!} \left(\frac{\beta}{1+\beta}\right)^{L \alpha}}
    {\sum_{L=0}^\infty \distPoisson(L \mid \overline{L}(\cX)) \, \distNegBinomial(1 \mid L\alpha, (1+\beta)^{-1})}  \cdot \alpha
    \\
    &= \frac{(1+\beta)^{-1} \sum_{L=1}^\infty \distPoisson(L \mid \overline{L}(\cX)) \, \frac{L!}{(L-1)!} \left(\frac{\beta}{1+\beta}\right)^{L \alpha}}
    {\sum_{L=0}^\infty \distPoisson(L \mid \overline{L}(\cX)) \, \frac{\Gamma(L\alpha + 1)}{1! \Gamma(L\alpha)} (1+\beta)^{-1} \left(\frac{\beta}{1 + \beta}\right)^{L \alpha} \bbI[L > 0]} \cdot \alpha
    \\
    &= \frac{(1+\beta)^{-1} \sum_{L=1}^\infty \distPoisson(L \mid \overline{L}(\cX)) \, L \, \left(\frac{\beta}{1+\beta}\right)^{L \alpha}}
    {(1+\beta)^{-1} \sum_{L=1}^\infty \distPoisson(L \mid \overline{L}(\cX)) \, L\alpha  \left(\frac{\beta}{1 + \beta}\right)^{L \alpha}}  \cdot \alpha \\
    &= 1.
\end{align}
This tedious algebra confirms the obvious fact that there is only one partition of the set~$\{1\}$, so the base case is trivially distributed as~$p(\cC \mid N)$.
The induction is complete.
\end{proof}

\begin{proof}[Proof of Lemma~\ref{lem:V_ratio}]
Substituting the definition \eqref{eq:V} and expanding the Poisson probability mass function,
\begin{align}
    \frac{V_{N,|\cC|+1}}{V_{N,|\cC|}} 
    &= \frac{\sum_{L=|\cC|+1}^\infty \frac{1}{L!} e^{-\overline{\eta}|\cX|} (\overline{\eta}|\cX|)^L \, \frac{L!}{(L-|\cC|-1)!} \left(\frac{\beta}{1+\beta}\right)^{L \alpha}}{\sum_{L=|\cC|}^\infty \frac{1}{L!} e^{-\overline{\eta}|\cX|} (\overline{\eta}|\cX|)^L \, \frac{L!}{(L-|\cC|)!} \left(\frac{\beta}{1+\beta}\right)^{L \alpha}} \\
    &= \frac{\sum_{L=|\cC|}^\infty (\overline{\eta}|\cX|)^{L+1} \, \frac{1}{(L-|\cC|)!} \left(\frac{\beta}{1+\beta}\right)^{(L+1) \alpha}}{\sum_{L=|\cC|}^\infty (\overline{\eta}|\cX|)^L \, \frac{1}{(L-|\cC|)!} \left(\frac{\beta}{1+\beta}\right)^{L \alpha}} \\
    &= \overline{\eta}|\cX| \left(\frac{\beta}{1 + \beta} \right)^{\alpha}.
\end{align}
\end{proof}

\begin{proof}[Proof of Theorem~\ref{thm:bkgd-partition-prior}]
The partition distribution changes slightly when we incorporate background data points. 
Intuitively, the background intensity is like a latent event that is always present.
Let $N_0$ denote the number of data points attributed to the background and let $\weight_0(\cX) = \int_{\cX} \lambda_0(x) \dif x$ denote the integrated background intensity.
\begin{align}
    p(N_0, N_1, \ldots, N_L \mid L)
    &= \distPoisson(N_0 \mid \weight_0(\cX)) \prod_{l=1}^L \int \distPoisson(N_l \mid \gamma_l) \, \mathrm{Ga}(\gamma_l \mid \alpha, \beta) \dif \gamma_l \\
    &= \distPoisson(N_0 \mid \weight_0(\cX)) \left( \frac{\beta}{1+\beta} \right)^{L\alpha} \left( \frac{1}{1+\beta} \right)^{N-N_0} \prod_{l=1}^L \frac{\Gamma(N_l + \alpha)}{N_l! \, \Gamma(\alpha)} .
\end{align}
There are~$\binom{N}{N_0,\ldots,N_L}$ parent assignments consistent with the latent cluster sizes~$N_0,\ldots,N_L$, and they are all equally likely under the prior, so the conditional probability of the parent assignments is,
\begin{align}
    p(\{z_n\}_{n=1}^N \mid L) 
    &= \binom{N}{N_0, \ldots, N_l}^{-1} \distPoisson(N_0 \mid \weight_0(\cX))  \,  \left( \frac{\beta}{1+\beta} \right)^{L\alpha} \left( \frac{1}{1+\beta} \right)^{N} \prod_{l=1}^L \frac{\Gamma(N_l + \alpha)}{N_l! \, \Gamma(\alpha)} \\
    &= \frac{1}{N!} e^{-\weight_0(\cX)} \weight_0(\cX)^{N_0} \left( \frac{\beta}{1+\beta} \right)^{L\alpha} \left( \frac{1}{1+\beta} \right)^{N-N_0} \prod_{l=1}^L \frac{\Gamma(N_l + \alpha)}{\Gamma(\alpha)}
\end{align}

The parent assignments induce a partition, but technically they assume an ordering of the latent events.  Moreover, if some of the latent events fail to produce any observed events, they will not be included in the partition.  In performing a change of variables from parent assignments~$\{z_n\}_{n=1}^N$ to partitions~$\cC$, we need to sum over latent event assignments that produce the same partition.  There are~${\binom{L}{|\cC|} |\cC|! = \frac{L!}{(L-|\cC|)!}}$ such assignments if there are~$L$ latent events but only~$|\cC|$ parts in the partitions. Thus,
\begin{align}
p(N, \cC_0, \cC \mid L)
&= \frac{L!}{(L-|\cC|)!} \frac{1}{N!} e^{-\weight_0(\cX)} \weight_0(\cX)^{|\cC_0|} \left( \frac{\beta}{1+\beta} \right)^{L\alpha} \left( \frac{1}{1+\beta} \right)^{N-|\cC_0|} \prod_{\cC_k \in \cC} \frac{\Gamma(|\cC_k| + \alpha)}{\Gamma(\alpha)}.
\end{align}
Of course,~$L$ must be at least~$|\cC|$ in order to produce the partition. 
Summing over the number of latent events~$L$ yields the marginal probability of the partition,
\begin{align}
p(N, \cC_0, \cC) &= \sum_{L=|\cC|}^\infty \distPoisson(L \mid \overline{L}(\cX)) \, p(N, \cC_0, \cC \mid L) \\
&= \left(\frac{(N-|\cC_0|)!}{N!} e^{-\weight_0(\cX)} \weight_0(\cX)(\cX)^{|\cC_0|}  \right) 
V_{N-|\cC_0|,|C|} \prod_{\cC_k \in \cC} \frac{\Gamma(|\cC_k|+\alpha)}{\Gamma(\alpha)},
\end{align}
where~$V_{N-|\cC_0|,|\cC|}$ is defined by eq.~\eqref{eq:V} above.
\end{proof}

\section{Random measure perspective on Neyman-Scott processes}
\label{sec:random-measures}
Let 
\begin{align}
    \label{eq:random-measure}
    G(\cA) &= \sum_{l=1}^L \weight_l \, \delta_{(\loc_l, \params_l)}(\cA)
\end{align}
denote the random discrete measure on~$\cX \times \Theta$ induced by a random sample of latent events from their Poisson process prior. 
(Here, $\delta_{(\loc,\params)}(\cA)$ is an indicator function that evaluates to 1 if $(\loc,\params) \in \cA$ and 0 otherwise.)
The random measure in eq.~\eqref{eq:random-measure} is the sum of functions applied to each point in a Poisson process, and as such it is a random variable. 
Under Assumption~\ref{asmp:gamma_weights}, it follows a Poisson-randomized gamma distribution defined by the following generative process,
\begin{align}
    L(\cA) &\sim \mathrm{Po}(\overline{L}(\cX) G_0(\cA)) \\
    G(\cA) &\sim \mathrm{Ga}(L(\cA) \alpha, \beta)    
\end{align}
where $L(\cA)$ is the number of latent events that fall in the set $\cA$ and
\begin{align}
    G_0(\cA) = \int_\cX \int_\Theta p(\loc, \params) \delta_{(\loc, \params)}(\cA) \dif \loc \dif \params
\end{align}
is the marginal probability of the set $\cA$; i.e. the \textit{base measure}. Marginalizing over the number of latent events, the expected measure scales linearly with the base measure $G_0(\cA)$,
\begin{align}
    \label{eq:G_mean}
    \E[G(\cA)] = \overline{L}(\cX) G_0(\cA) \alpha / \beta.
\end{align} 
Let $(\cA_1, \ldots, \cA_M)$ be a partition of $\cX$. The random measures $(G(\cA_1), \ldots, G(\cA_M))$ are independent Poisson-randomized gamma variables with means given by~\cref{eq:G_mean}.

By Campbell's Theorem~\citep[Ch. 3.2]{kingman1992poisson}, the moment generating function of $G(\cA)$ is,
\begin{align}
    \E[e^{-t G(\cA)}] 
    &= \exp \left\{ \int_\cX \int_{\reals_+} \int_\Theta \latentrate(\loc, \weight, \params) \left(e^{-t \weight \delta_{(\loc, \params)}(\cA)} - 1 \right) \dif \loc \dif \weight \dif \params  \right\} \\
    &= \exp \left\{ \overline{L}(\cX) G_0(\cA) \int_{\reals_+} \distGamma(\weight \mid \alpha, \beta) \left(e^{-t \weight} - 1 \right) \dif \weight \right\}.
\end{align}
Taking the limit described in Corollary~\ref{cor:nsp_dpmm}, we have,
\begin{align}
    \lim_{\substack{\alpha \to 0 \\ \alpha \overline{L}(\cX) \to \gamma}} \E[e^{-t G(\cA)}] 
    &= 
    \!\!\lim_{\substack{\alpha \to 0 \\ \alpha \overline{L}(\cX) \to \gamma}} \!\!
    \exp \left\{\overline{L}(\cX) G_0(\cA) \int_{\reals_+} \frac{\alpha \beta^\alpha }{\alpha \Gamma(\alpha)} \weight^{\alpha-1} e^{-\beta \weight} \left(e^{-t \weight} - 1 \right) \dif \weight \right\} \\
    &= 
    \exp \left\{ \gamma G_0(\cA) \int_{\reals_+} \weight^{-1} e^{-\beta \weight} \left(e^{-t \weight} - 1 \right) \dif \weight \right\} \\
    &= 
    \label{eq:gamma_mgf}
    \left(1 + \frac{t}{\beta} \right)^{-\gamma G_0(\cA)},
\end{align}
which follows from the fact that $\Gamma(\alpha+1) = \alpha \Gamma(\alpha)$ and $\Gamma(1) = 1$, and from the L\'{e}vy-Khinchine representation of the gamma distribution. 
Eq.~\eqref{eq:gamma_mgf} is the MGF of a gamma distribution, $G(\cA) \sim \distGamma(\gamma G_0(\cA), \beta)$, so in this limit the random measure is a gamma process.
Intuitively, the Neyman-Scott process with gamma weights approaches a gamma process in the limit where there are infinitely many latent events with weights going to zero. 
When the gamma process is normalized by its total mass it yields a Dirichlet process. 

\section{Inference Details}
\label{app:inference}

\sloppy
\subsection{Sampling the background intensity}
After sampling the partition, we update the background intensity given the data points assigned to the background cluster,~${\{x_n: n \in \cC_0\}}$. 
We assume access to MCMC transition operators that leave the conditional distribution,~$p(\lambda_0(x) \mid \{x_n: n \in \cC_0\})$, invariant. 
For the simple, homogenous background intensity model with a gamma prior~$\overline{\lambda}_0 \sim \distGamma(\alpha_0, \beta_0)$, the conditional distribution of~$\overline{\lambda}_0$ is available in closed form,
\begin{align}
    p(\overline{\lambda}_0 \mid \{x_n: n \in \cC_0\}) 
    &= \distGamma(\overline{\lambda}_0 \mid \alpha_0 + |\cC_0|, \beta_0 + |\cX|).
\end{align}

\subsection{Sampling the latent events} 
Given the partition and global parameters, it is straightforward to sample the latent events.  
We update the latent event parameters by sampling $p(\loc_k, \params_k \mid \{x_n: n \in \cC_k\})$ for each cluster. 
If~$p(\loc, \params)$ in the latent event intensity has any parameters, they can be updated given the samples~$\{(\loc_k, \params_k)\}_{k=1}^{|\cC|}$. 
In our three examples,~$p(\loc, \params)$ is an exponential family density, so we place a conjugate prior on its parameters to enable simple Gibbs updates. 
Otherwise, the parameter estimates are only for visualization purposes; they are immediately discarded before the subsequent partition updates, which marginalize over parameters of the clusters. 

\subsection{Sampling the total number of latent events} 
Finally, we have at least two choices when it comes to updating the hyperparameters~$\alpha$, $\beta$, and the latent event intensity~$\overline{L}(\cX)$.
First, we can place a prior distribution over them and sample their conditional distribution.
Conditional sampling is most easily accomplished by introducing the total number of latent events as an auxiliary variable (recall that the number of latent events was collapsed out in the derivation of the partition distribution). 
The following lemma gives the necessary conditional distribution.
\begin{lemma}
In a Neyman-Scott process with gamma weights (Assumption~\ref{asmp:gamma_weights}), the number of \emph{empty} clusters --- i.e. the number of latent events that produce zero observed data points --- is a Poisson random variable that is independent of the number of occupied clusters,
\begin{align}
    L - |\cC| &\sim \distPoisson\left( \overline{L}(\cX) \left(\frac{\beta}{1+\beta} \right)^\alpha \right).
\end{align}
\end{lemma}
\begin{proof}
We have,
\begin{align}
    p(L \mid N, \cC) 
    &\propto p(N, \cC \mid L) \, p(L) \\
    &\propto \frac{1}{(L - |\cC|)!} \left(\frac{\beta}{1+\beta}\right)^{L\alpha} \overline{L}(\cX)^L \\
    &\propto \mathrm{Po}\left(E \;\Big|\; \overline{L}(\cX) \left(\frac{\beta}{1+\beta} \right)^\alpha \right).
\end{align}
where~$E = L - |\cC|$ is the number of empty clusters and~$p(N, \cC \mid L)$ is given by~\cref{eq:p_NC_given_L}.
\end{proof}
Interestingly, the mean of this conditional distribution is the same as the ratio in Lemma~\ref{lem:V_ratio}. 

\subsection{Sampling the latent event rate}
Many Neyman-Scott processes, including the three examples above, parameterize the total measure of the latent event intensity as a linear function of the volume,~$\overline{L}(\cX) = \overline{\latentrate}|\cX|$.
Under a conjugate gamma prior~$\overline{\latentrate} \sim \distGamma(\alpha_\latentrate, \beta_\latentrate)$, its conditional distribution is,
\begin{align}
    p(\overline{\latentrate} \mid L) &= \distGamma(\overline{\latentrate} \mid \alpha_\latentrate + L, \beta_\latentrate + |\cX|).
\end{align}

\subsection{Sampling the parameters of the weight distributions} 
Similarly, we can introduce auxiliary weights for the empty clusters from their gamma conditional distribution,
\begin{align}
    \weight_l &\sim \distGamma(\alpha, \beta + 1) \quad \text{for } l=|\cC|+1, \ldots,L
\end{align}
and then sample the hyperparameters~$\alpha$ and~$\beta$ from their conditional distribution,
\begin{align}
    p(\alpha, \beta \mid \{\weight_l\}_{l=1}^L) &\propto
    p(\alpha, \beta) \prod_{l=1}^L \distGamma(\weight_l \mid \alpha, \beta).
\end{align}
If the prior factors as $p(\alpha, \beta) = p(\alpha) \, \distGamma(\beta \mid \alpha_\beta, \beta_\beta)$, the conditional distribution of the inverse scale is,
\begin{align}
    p(\beta \mid \{\weight_l\}_{l=1}^L) &= 
    \distGamma \left(\beta \mid \alpha_\beta + L\alpha, \beta_\beta + \sum_{l=1}^L \weight_l \right).
\end{align}
The shape parameter~$\alpha$ does not have a simple conjugate prior, but is amenable to Metropolis-Hastings updates. 
However, we prefer to treat~$\alpha$ as a hyperparameter and select it via cross-validation.

\subsection{Split-merge moves}
\label{app:split-merge}

\citet{Miller2018} noted that their collapsed Gibbs sampler for MFMMs was substantially improved by the addition of split-merge moves~\citep{Jain2007}, which also act directly on the partition.
The closed-form partition probability function for the NSP (\Cref{thm:partition-prior}) enables split-merge moves as well.
We find that split-merge moves combined with collapsed Gibbs updates are an effective combination for posterior sampling.

Split-merge moves are especially important when the prior on weights,~$\gamma_\ell \sim {\mathrm{Ga}(\alpha, \beta)}$, places little mass of low-weight latent events.
In that regime, the sampler is unlikely to create new latent events and is therefore slow to explore different partitions of observed events.
Conversely, when a latent event is the parent of many data points, the sampler is unlikely to eliminate it, since the event must pass temporarily pass through a low-probability phase where it is the parent of only a few assigned data points.
This problem is common to other nonparametric Bayesian mixture models as well \cite[e.g.]{Miller2018}.
If, on the other hand, the variance of~$\mathrm{Ga}(\alpha, \beta)$ is large relative to the mean, then the probability of forming new clusters or eliminating existing clusters is non-negligible and the sampler tends to mix more effectively.
Unfortunately, this latter regime is also probably of lesser scientific interest, since the interesting latent events are typically large in amplitude. 
For example, in neuroscience settings, sequences of interest may involve many thousands of observed events, each potentially contributing a small number of spikes~\cite{Buzsaki2018,Hahnloser2002}.

Split-merge moves~\citep{Jain2007} are an effective means of addressing this issue.
In a split move, we choose an existing cluster $\cC_k$ and propose to split it into two new clusters $\hat \cC_1, \hat \cC_2$ (such that $\hat \cC_1 \cup \hat \cC_2 = \cC_k$ and $\hat \cC_1 \cap \hat \cC_2 = \emptyset$).
To choose $\hat \cC_1, \hat \cC_2$, we first randomly assign the elements of $\cC_k$ among the two clusters with equal probability, then run a few Gibbs updates restricted to the two clusters.
In a merge move, we choose two existing clusters $\cC_k, \cC_{k'}$ and propose merging them into a single cluster $\hat \cC = \cC_k \cup \cC_{k'}$.
In both moves, we accept the proposal with probability defined by the Metropolis-Hastings acceptance ratio.
Like the collapsed Gibbs updates, split-merge moves operate directly on the posterior over partitions, marginalizing over the cluster locations and parameters.
These moves thus require our derivation of the posterior over partitions and are a simple extension of the collapsed Gibbs updates.
For algorithm psuedo-code, please refer to~\citet{Jain2007}.

\subsection{Initialization}
\label{app:initialization}
In our synthetic experiments without a background intensity, we initialize each Markov chain by sampling the $N$ parent assignments uniformly at random from the set $\{1, \ldots, N\}$.
Since there are at most $N$ \textit{observed} clusters, this initialization procedure assigns non-zero probability,
\begin{align*}
    p_0(\cC) = \frac{N!}{(N - |\cC|)!} \cdot \frac{1}{N^N}
\end{align*}
to each possible partition.
For Neyman-Scott processes with a background intensity, we initially assign each parent assignment to the background cluster with probability $q$ and otherwise sample a parent assignment uniformly at random from $\{1, \ldots, N\}$ with probability $1-q$.
When using samplers such as RJMCMC, we also sample cluster parameters conditioned on the initial partition.

In our neuroscience and document stream experiments, we use an an annealing procedure to initialize the Markov chains.
We fix the mean of the cluster weights~$\gamma_\ell \sim \mathrm{Ga}(\alpha, \beta)$ and adjust~$\alpha$ and~$\beta$ to gradually decrease the variance of $\gamma_\ell$ from some high value down to the variance specified by the prior.
Initially, the sampler produces many small clusters of spikes, and as we lower the variance of~$\mathrm{Ga}(\alpha, \beta)$ to a target value, the Markov chain typically combines these clusters into larger sequences.
Finally, though we have not found it necessary, one could use alternative methods, like convolutional matrix factorization~\citep{Smaragdis2006, degleris2019fast} to initialize the MCMC algorithm.

\subsection{Parallel MCMC}
Resampling the partition (i.e. cluster assignments) is the primary computational bottleneck for the Gibbs sampler.
For many datasets, we can improve performance substantially, at the cost of minor approximation error, by parallelizing the computation~\cite{Angelino2016}.
Consider a set of observed events on the time interval $[0,T]$. Given~$P$ processors, we divide the dataset into intervals lasting~$T / P$ seconds, and allocate one interval per processor.
The current global parameters,~$\Theta$, are first broadcast to all processors.
In parallel, the processors update the cluster assignments for the observed events in their interval, and then send back sufficient statistics describing each inferred cluster within their interval.
After these sufficient statistics are collected on a single processor, the global parameters are re-sampled and then broadcast back to the processors to initiate another iteration.
This algorithm introduces some error since clusters are not shared across processors.
In essence, this introduces erroneous edge effects if a cluster is split across two processors.
However, these errors are negligible when the cluster duration is much less than~$T / P$, which we expect is the practical regime of interest.

\subsection{Hyperparameter selection by cross-validation}

\begin{figure}[t]
\centering
\includegraphics[width=\linewidth]{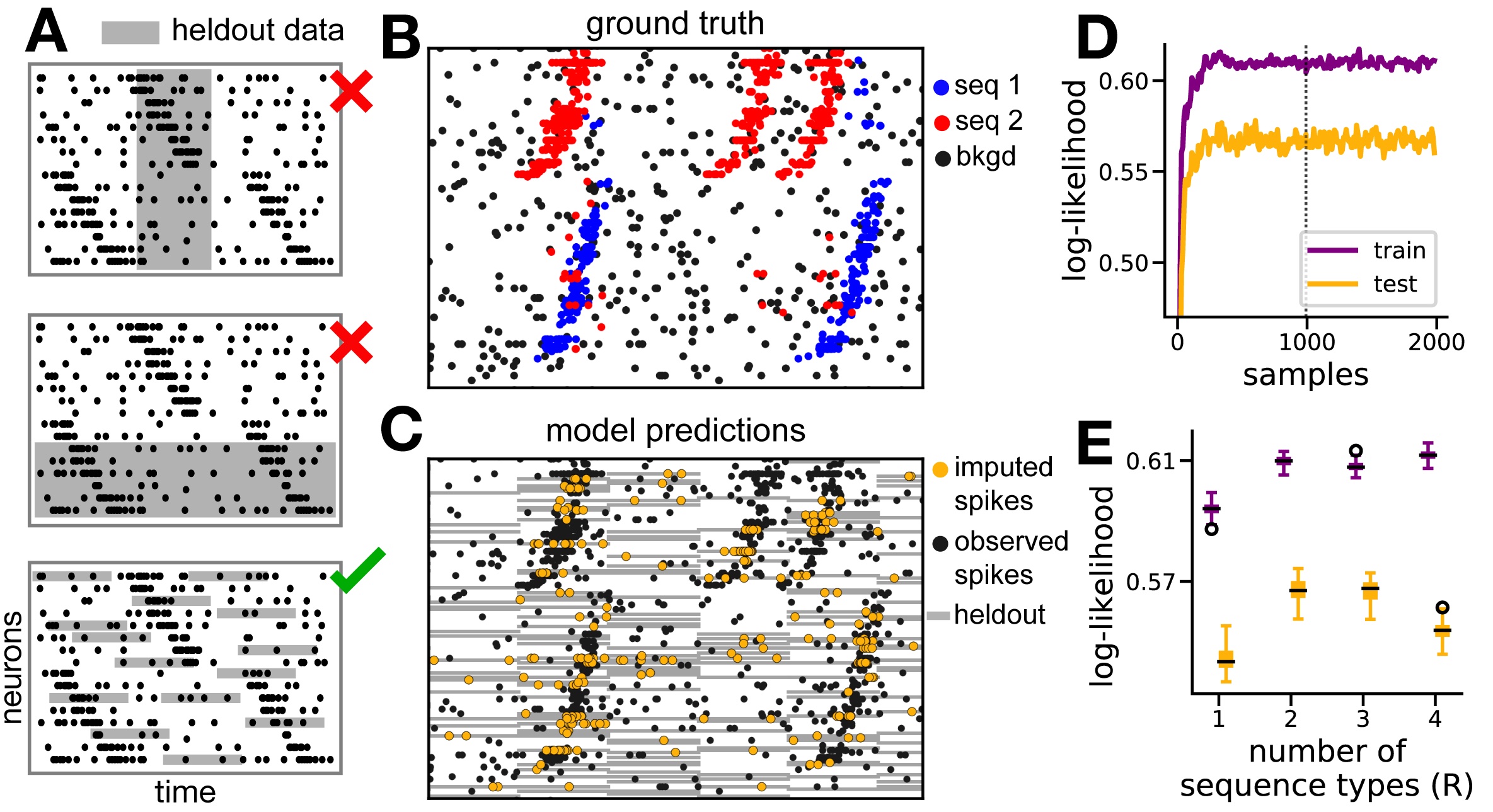}
\caption{
\textit{(A)} Schematic of train/test partitions. We propose a speckled holdout pattern (bottom).
\textit{(B)} A subset of a synthetic spike train containing two sequences types.
\textit{(C)} Same data, but with grey regions showing the censored test set and yellow dots denoting imputed spikes.
\textit{(D)} Log-likelihood over Gibbs samples; positive values denote excess nats per unit time relative to a homogeneous Poisson process baseline.
\textit{(E)} Box plots showing range of log-likelihoods on the train and test sets for different choices of~$R$; cross-validation favors~$R=2$, in agreement with the ground truth shown in panel B.
}
\label{fig:crossval}
\end{figure}

In a fully Bayesian analysis, the hyperparameters could be given weak priors and sampled within the MCMC loop, as described in the optional steps of Algorithm~\ref{alg:collapsedgibbs}.
Alternatively, a practitioner could use cross-validation to estimate hyperparameters based on the log probability assigned to held-out data.
Partitioning the data into training and testing sets must be done somewhat carefully.
For example, in the neural sequence detection example, we cannot withhold time intervals completely or else the model will not accurately predict latent sequences occurring in these intervals; likewise, we cannot withhold individual neurons completely or else the model will not accurately predict the response parameters of those held out cells.
Thus, we adopt a ``speckled'' holdout strategy \cite{Wold1978}. See~\cref{fig:crossval}A.
We treat held-out data points as missing data and sample them as part of the MCMC algorithm.  (Their conditional distribution is fully specified by the generative model.)

\Cref{fig:crossval}B-C illustrate this missing data imputation procedure synthetic data from the neural spike train model described in~\cref{sec:spike_trains}. Here, there are two sequence types (red and blue). The gray regions in~\cref{fig:crossval}C denote held-out neurons and intervals, and the yellow dots are imputed spikes from one sample of the MCMC chain. \Cref{fig:crossval}D shows the likelihood of the train and test (held-out) spikes as a function of MCMC iteration: both plateau after a few hundred iterations. \Cref{fig:crossval}E shows how the log likelihood of test spikes can be used to set hyperparameters, like the number of sequence types, using cross-validation. Here, the test log likelihood peaks at the true number of sequence types. 


\subsection{Reversible-Jump MCMC}

For comparison, a standard approach to inference in Neyman-Scott processes is reversible-jump MCMC (RJMCMC). 
The reversible jump MCMC sampler alternates between proposing latent event birth-death moves, sampling parent assignments conditioned on the latent event parameters, and sampling of the latent event parameters conditioned on parent assignments.

\paragraph{Birth-death moves} 
Let $q_{\mathsf{birth}}$ denote the probability of proposing a ``birth'' move. (The probability of proposing a ``death'' move is $1-q_{\mathsf{birth}}$.) In a birth move, a new latent event is added with location, weight, and parameters randomly drawn from the prior, $\loc_{L+1}, \weight_{L+1}, \params_{L+1} \sim p(\loc, \weight, \params)$. In a death move, one of the $L$ latent events is deleted at random. Since the new events are drawn from the prior, many terms cancel in the MH acceptance probability. 

The probability of accepting a birth move is,
\begin{align}
    a_{\mathsf{birth}} &= \min \left\{1, \; \frac{1 - q_{\mathsf{birth}}}{q_{\mathsf{birth}}} \frac{\overline{L}(\cX)}{L+1} \frac{p(\{(x_n, y_n)\}_{n=1}^N \mid \{(\loc_l, \weight_l, \params_l)\}_{l=1}^{L+1})}{p(\{(x_n, y_n)\}_{n=1}^N \mid \{(\loc_l, \weight_l, \params_l)\}_{l=1}^{L})} \right\},
\end{align}
where ${p(\{(x_n, y_n)\}_{n=1}^N \mid \{(\loc_l, \weight_l, \params_l)\}_{l=1}^{L+1})}$ is the Poisson process likelihood function,
\begin{multline}
    p(\{(x_n, y_n)\}_{n=1}^N \mid \{(\loc_l, \weight_l, \params_l)\}_{l=1}^L) \\
    = \exp \left\{ -\int \lambda\left(x, y \mid \{(\loc_l, \weight_l, \params_l)\}_{l=1}^L \right) \dif x \dif y \right\}
    \prod_{n=1}^N \lambda \left(x_n, y_n \mid \{(\loc_l, \weight_l, \params_l)\}_{l=1}^L \right),
\end{multline}
and the intensity is given by~\cref{eq:nsp_intensity_1}. 
Likewise, the probability of accepting a death move that deletes the $i$-th latent event is,
\begin{align}
    a_{\mathsf{death}} &= \min \left\{1, \; \frac{q_{\mathsf{birth}}}{1 - q_{\mathsf{birth}}} \frac{L}{\overline{L}(\cX)} \frac{p(\{(x_n, y_n)\}_{n=1}^N \mid \{(\loc_l, \weight_l, \params_l)\}_{l=1}^L \setminus (\loc_i, \weight_i, \params_i))}{p(\{(x_n, y_n)\}_{n=1}^N \mid \{(\loc_l, \weight_l, \params_l)\}_{l=1}^{L})} \right\}.
\end{align}
Note that if the impulse responses are concentrated on small regions around the latent events, the likelihood ratios can be computed more efficiently by ignoring data points $(x_n, y_n)$ in the product that are far from the proposed latent event.

\paragraph{Parent and parameter updates} 
We combine the birth-death moves with Gibbs updates that introduce auxiliary ``parent'' variables, which assign observed data points to one of the latent events based on the relative intensity of the impulse responses. This follows from the additive nature of the Neyman-Scott process and the Poisson superposition principle (see~\cref{sec:simulating}). Once data points have been assigned to latent events, we perform a Gibbs update of the latent event locations, weights, and parameters, leveraging to the assumption that the model is conditionally conjugate. 

Thus, the RJMCMC algorithm benefits from the conjugacy of the model, just like collapsed Gibbs algorithm. However, it does not benefit from the closed-form partition distribution derived in this work, which enables direct sampling of the partition distribution via collapsed Gibbs and split-merge moves.

\subsection{Performance of CG and RJMCMC as a function of dimension}

\begin{figure}[t]
\centering
\includegraphics[width=\linewidth]{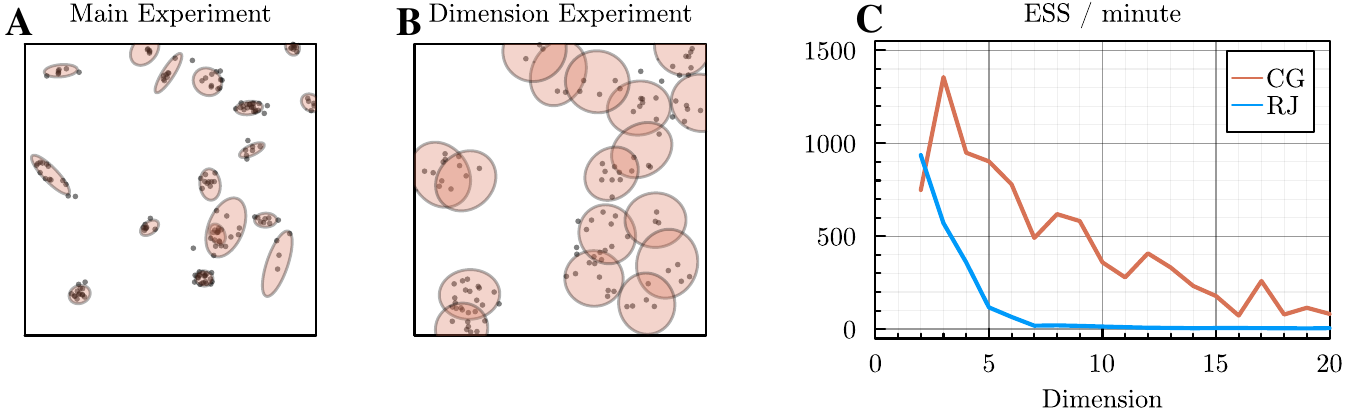}
\caption{%
\textit{Performance of the collapsed Gibbs (CG) and reversible jump MCMC (RJ) samplers as a function of problem dimension.}
\textbf{(A)} Sample data used to compare CG to RJ in Figure~\ref{fig:cg-rj}.
\textbf{(B)} Two-dimensional sample data used to compare CG to RJ in Panel~C.
\textbf{(C)} Effective sample size per minute as a function of data dimension.
}
\label{fig:dimension}
\end{figure}

To better understand the performance of the collapsed Gibbs sampler with split-merge moves, we compare its performance with that of reversible jump MCMC on datasets of different dimensions.
Specifically, using the same experimental setup as in Section~\ref{sec:sampler-experiment}, we vary the dimension $D \in \{2, \ldots, 20\}$ of the data and measure the performance of each sampler.
To mitigate the effect of dimension on the distribution over each cluster covariance $\theta \sim \mathrm{IW}(\cdot \mid d, \Psi)$, we set the degrees of freedom $d$ to a large constant (e.g., $d=100$).
We also increase the average within-cluster variance so that the clusters occupy a large fraction of the problem volume when $D = 2$.
Figures~\ref{fig:dimension}A and~\ref{fig:dimension}B display sample data from the experiment in Section~\ref{sec:sampler-experiment} and the current experiment, respectively, to illustrate these changes.

Figure~\ref{fig:dimension}C shows the effective sample size of each sampler after 60 seconds as a function of dimension.
In the two dimensional setting, the reversible jump sampler outperforms the collapsed Gibbs sampler.
Intuitively, this occurs because the cluster occupy a large fraction of the total problem volume, so the reversible jump sampler's cluster proposals are likely to be accepted.
However, as dimension increases, the performance of the reversible jump sampler rapidly decays to zero.
This is because the volume of each cluster (measured by the 2-$\sigma$ covariance ellipse, for example) decreases exponentially relative to the volume of $[0, 1]^D$ as the dimension $D$ increases, so most random cluster proposals will be rejected.
In contrast, the collapsed Gibbs algorithm samples directly from the posterior over partitions and does not suffer from this issue to the same extent.
These results suggest our collapsed Gibbs algorithm is particularly useful in high dimensional settings, such as the document stream modeling experiment in Section~\ref{subsec:document-model}.
\section{Experimental Details}
\label{app:expm_supp}

This appendix includes further details on the experiments. 

\paragraph{Computing infrastructure}
All experiments were performed on a 2017 MacBook Pro (3.1 GHz Intel Core i7, 4 cores, 16 GB RAM). 

\subsection{Synthetic data experiments}
For the collapsed Gibbs with split-merge vs RJMCMC experiments in~\Cref{fig:cg-rj} and the NSP vs DPMM experiments in~\Cref{fig:nsp-dpm}, we generated synthetic data by simulating an NSP on the unit square~$\cX = [0,1]^2$ without background rate using the following priors,
\begin{itemize}
    \item Expected number of latent events: $\overline{L}(\cX) = 20.0$ 
    \item Latent event weight concentration: $\alpha = 9.0$
    \item Latent event weight rate: $\beta = 0.3$
    \item Latent event location prior: $\loc_l \sim \mathrm{Unif}([0,1]^2)$
    \item Latent event covariance prior: $\params_l \sim \mathrm{IW}(5.0, 10^{-3} I)$
\end{itemize}

For each MCMC algorithm (RJMCMC or CG+SM), we ran three independent chains for 2 minutes ($\approx 7,000$ iterations) each, starting from the random initializations as described in~\cref{app:initialization}. For CG+SM, we performed 10 split-merge moves per collapsed Gibbs sweep. For RJMCMC, we performed 10 birth-death moves per parent and parameter Gibbs update. Birth and death moves were equally probable ($q_{\mathsf{birth}} = 0.5$.) Potential scale reduction factor (PSR), aka $\hat{R}$, and effective sample size (ESS) were computed using the \texttt{MCMCDiagnosticTools.jl} package from \url{https://github.com/TuringLang/MCMCDiagnosticTools.jl}. Plots are shown as a function of wall-clock time 

\subsection{Sequence detection in neural spike train experiments}

For the sequence detection in neural spike train experiments presented in~\Cref{fig:bird_fig}, the domain is the time interval $\cX = [0, T]$ and the marks are integers $y_n \in \{1,\ldots, Y\}$. We used the following hyperparameter settings,
\begin{itemize}
    \item Number of sequence types: $S=2$
    \item Latent event rate: $\overline{\nu} = 1$ (so expected number of latent events is $\overline{L}(\cX) = \overline{\nu} T$)
    \item Latent event weight concentration: $\alpha = 10.0$
    \item Latent event weight rate: $\beta = 0.1$
\end{itemize}

We placed the following weak priors on the key model parameters and sampled them inside the MCMC loop, as described in~\Cref{alg:collapsedgibbs},
\begin{itemize}
    \item Background rate prior: $\overline{\lambda}_0 \sim \mathrm{Ga}(30.0, 1.0)$
    \item Sequence type distribution prior: $\pi \sim \mathrm{Dir}([1, 1])$
    \item Background neuron distribution prior: $a_0 \sim \mathrm{Dir}(0.3 \cdot \boldsymbol{1}_Y)$
    \item Sequence neuron distribution prior: $a_s \sim \mathrm{Dir}(0.1 \cdot \boldsymbol{1}_Y)$
    \item Sequence width prior: $\sigma_{y,s}^2 \sim \mathrm{Inv}-\chi^{2}(1.0, 0.5)$ (a scaled inverse chi-squared prior)
    \item Sequence latency prior: $b_{y,s} \mid \sigma_{y,s}^2 \sim \cN(0, \sigma_{y,s}^2)$
\end{itemize}

We initialized the sampler with a simple annealing procedure as described in~\citet{Williams2020_ppseq}. (We expect that the initialization procedure described in~\Cref{app:initialization} would work as well.) We ran 3 independent MCMC chains for 2000 iterations each, using 10 split-merge moves per collapsed Gibbs sweep. We visually assessed the converge of the MCMC algorithm by inspecting the log probability trace. Posterior distributions shown in~\Cref{fig:bird_fig}B-E were estimated using the last 1000 samples.

\subsection{Event detection in diplomatic cables experiments}

For the event detection example from~\Cref{subsec:document-model}, we used the following hyperparameters,
\begin{itemize}
    \item Latent event rate: $\overline{\nu} = \tfrac{1}{30}$ [one latent event per month in expectation]
    \item Latent event weight concentration: $\alpha = 2500.0$
    \item Latent event weight rate: $\beta = 5.0$
    \item Latent event width: $\sigma^2 = 2.0$ [days]
    \item Author concentration: $\alpha_a = \boldsymbol{1}_A$
    \item Word rate prior: $\alpha_c = 1$, $\beta_c = 1$
\end{itemize}
These hyperparmeters were set such that the gamma prior on latent event weights had mean $500$ and variance $100$. Likewise, the word rate prior is such that the prior over \emph{normalized} word distribution $[\theta_1^{(c)}, \ldots, \theta_V^{(c)}] / \sum_v \theta_v^{(c)}$ is uniform on $\Delta_V$.

We placed the following priors on the model parameters and sampled them inside the MCMC loop, as described in~\Cref{alg:collapsedgibbs},
\begin{itemize}
    \item Background rate prior: $\overline{\lambda}_0 \sim \mathrm{Ga}(10^5, 10)$
\end{itemize}
These hyperparmeters were set such that the gamma prior on the background rate had mean $1000$ and variance $100$.

We set the background author distribution and word rates using a heuristic similar to empirical Bayes. We set $\theta_0^{(a)}$ equal to the empirical distribution of authors and $\phi_{a,v}$ proportional to the frequency with which author $a$ used word $v$ in the data. 

The baseline model can be viewed as an NSP with a fixed set of latent events, one per week. We use the same hyperparameters and prior distributions as above, except we do not have a latent event rate since the number of latent events is fixed. We perform MCMC using the same algorithm, except we do not update latent event times during sampling.

Given the complexity of running this MCMC algorithm on a dataset with $N=34,732$ cables and a $V=21,819$ word vocabulary, we only ran the MCMC algorithm for 300 iterations. The log probability appears to have stabilized at this point, but we cannot make strong claims of convergence. Nevertheless, the resulting partition of events recovers clear world events.
\section{Additional Proofs}
\label{app:add-proofs}

This appendix provides additional theorems and proofs about the Neyman-Scott process, its limiting relationship to the Dirichlet process, and its self consistency.

\subsection{Posterior distribution over partitions under the NSP and DPMM}

\Cref{sec:random-measures} provided a random measure perspective on the Neyman-Scott process and connected it to the gamma and Dirichlet processes. Here we show a more direct equivalence between the posterior distribution over partitions (and hence number of clusters) under a DPMM and an NSP in the limit described in \Cref{cor:nsp_dpmm}.

\begin{theorem}
In the limit that~$\alpha \to 0$ while~$\alpha \, \overline{L}(\cX) \to \gamma$ (like in \Cref{cor:nsp_dpmm}), the posterior distribution over partitions (and number of clusters) under an NSP with gamma weights coincides with the posterior under a DPMM with concentration~$\gamma$ and a base measure with density~$p(\loc, \params)$.
\end{theorem}

\begin{proof}
In this limit, the expected number of latent events, $\overline{L}(\cX)$, must go to infinity while the individual latent event weights go to zero. We begin by showing the posterior of the partition over $N$ data points for an NSP converges to that of DPMM, given a fixed dataset of $N$ data points.

The posterior of the partition for NSP given $N$ data points $\{(x_n, y_n)\}_{n=1}^N$ is
\begin{align}
p(\cC \mid \{(x_n, y_n)\}_{n=1}^N)
& \propto p(\cC \mid N) \, p(\{(x_n, y_n)\}_{n=1}^N \mid \cC)
\end{align}
Since both DPMM and NSP share the same likelihood function, $p(\{(x_n, y_n)\}_{n=1}^N \mid \cC)$, it is sufficient to establish the convergence for only the prior, $p(\cC \mid N)$. 

The prior partition distribution under an NSP (\Cref{thm:partition-prior}) is 
\begin{align}
    p(\cC \mid N) &\propto p(\cC, N) 
    = V_{N,|\cC|} \prod_{\cC_k \in \cC} \frac{\Gamma(|\cC_k| + \alpha)}{\Gamma(\alpha)}\\
    &= \left[\frac{1}{N!} \left(\frac{1}{1+\beta} \right)^{N} \sum_{L=|\cC|}^\infty \distPoisson(L \mid \overline{L}(\cX)) \, \frac{L!}{(L-|\cC|)!} \, \left(\frac{\beta}{1+\beta}\right)^{L \alpha} \right] \prod_{\cC_k \in \cC} \frac{\Gamma(|\cC_k| + \alpha)}{\Gamma(\alpha)}\\
    &\propto\left(\frac{1}{\Gamma(\alpha)}\right)^{|\cC|} \Bigg[ \sum_{L=|\cC|}^\infty \frac{e^{- \overline{L}(\cX)}}{(L-|\cC|)!} \left(\overline{L}(\cX) \left(\frac{\beta}{1+\beta}\right)^\alpha \right)^L \Bigg] \prod_{\cC_k \in \cC} \Gamma(|\cC_k| + \alpha) \\
    &\propto\left(\frac{\overline{L}(\cX) \left(\frac{\beta}{1+\beta} \right)^\alpha}{\Gamma(\alpha)}\right)^{|\cC|} \left[ \sum_{L=0}^\infty \frac{e^{- \overline{L}(\cX)}}{L!} \, \left(\overline{L}(\cX) \left(\frac{\beta}{1+\beta}\right)^\alpha \right)^L \right] \prod_{\cC_k \in \cC} \Gamma(|\cC_k| + \alpha) \\
    &= \left(\frac{\alpha \overline{L}(\cX) \left(\frac{\beta}{1+\beta} \right)^\alpha}{\Gamma(\alpha + 1)}\right)^{|\cC|} \left[ \sum_{L=0}^\infty \frac{e^{- \overline{L}(\cX)}}{L!} \, \left(\overline{L}(\cX) \left(\frac{\beta}{1+\beta}\right)^\alpha \right)^L \right] \prod_{\cC_k \in \cC} \Gamma(|\cC_k| + \alpha)
\end{align}

Now take the limit as~$\alpha \to 0$ while~$\alpha \, \overline{L}(\cX) \to \gamma$. Then $(\frac{\beta}{1+\beta})^\alpha \to 1$ and the term inside the sum reduces to Poisson pmf,~$\mathrm{Po}(L \mid \overline{L}(\cX))$. The sum of the Poisson pmf converges to 1 even in the limit as the rate goes to infinity. In that limit, 
\begin{align}
    \!\!\lim_{\substack{\alpha \to 0 \\ \alpha \overline{L}(\cX) \to \gamma}} \!\! p(\cC \mid N) 
    &\propto  \gamma^{|\cC|}  \prod_{\cC_k \in \cC} \Gamma(|\cC_k|).
\end{align}
This limit of the prior partition distribution coincides with that of the DPMM with concentration parameter $\gamma$, i.e.  $p_{\mathrm{DPMM}}(\mathcal{C} \mid N) = \frac{\gamma^{|\cC|}}{\gamma^N}\prod_{\cC_k\in \cC} \Gamma(|\cC_k|)$. We thus have established the convergence between the partition distribution of NSP and that of DPMM.

The convergence between the partition distribution implies that the posterior of the number of clusters in the partition for NSP also converges to that of DPMM because
$p(|\cC|= t \mid \{(x_n, y_n)\}_{n=1}^N) = \sum_{\cC:|\cC|=t} p(\cC \mid \{(x_n, y_n)\}_{n=1}^N)$.

\end{proof}









\subsection{Self-consistency of the Neyman-Scott process}

Let $p(\cC \mid N)$ denote the distribution on partitions of $[N] = \{1, \ldots, N\}$ induced by the Neyman-Scott process, as defined in \Cref{thm:urn_process}. 

\begin{proposition}
\label{prop:self-consistent}
The marginal distribution on partitions of $[M]$ obtained by sampling $p(\cC \mid N)$ and discarding indices $M+1,\ldots,N$ (assuming $M < N$) is $p(\cC \mid M)$. In other words, the NSP has self-consistent marginals.
\end{proposition}

We can see this from the generative process for partitions. A partition of $[N] = \{1,\ldots,N\}$ can be sampled by iteratively applying the transition operator from Theorem 2. Removing integers $M+1,\ldots,N$ for $M < N$ yields a partition distributed as $p(\cC \mid M)$, since the distribution of the partition of $[M]$ is unaffected by later indices. 

Kolmogorov's extension theorem implies the existence of a unique probability distribution on partitions of the positive integers $\mathbb{N}$ such that the marginal distribution of partitions of $[N]$ is $p(\cC \mid N)$ for all $N \in \mathbb{N}$. This proposition is analogous to Proposition 3.3 of~\citet{Miller2018}. 

\singlespacing

\bibliography{refs}
\bibliographystyle{unsrtnat}

\end{document}